\documentclass{IEEEtran}
\usepackage{amssymb, amsmath, amsthm}
\usepackage{bm}
\usepackage{color}
\usepackage{graphicx}
\usepackage{soul}
\usepackage{hyperref}
\usepackage{enumerate}
\usepackage{algorithm, algorithmic}

\graphicspath{{figures/}}

\newtheorem{definition}{Definition}
\newtheorem{proposition}{Proposition}

\newcommand{\bfgamma}{\boldsymbol{\gamma}}

\newcommand{\bftau}{\boldsymbol{\tau}}

\newcommand{\bfomega}{\boldsymbol{\omega}}

\newcommand{\bfb}{{\bm{b}}}

\newcommand{\bfd}{{\bm{d}}}
\newcommand{\bfe}{{\bm{e}}}
\newcommand{\bff}{{\bm{f}}}
\newcommand{\bfg}{{\bm{g}}}

\newcommand{\bfn}{{\bm{n}}}

\newcommand{\bfp}{{\bm{p}}}
\newcommand{\bfq}{{\bm{q}}}
\newcommand{\bfr}{{\bm{r}}}

\newcommand{\bft}{{\bm{t}}}
\newcommand{\bfu}{{\bm{u}}}
\newcommand{\bfv}{{\bm{v}}}
\newcommand{\bfw}{{\bm{w}}}
\newcommand{\bfx}{{\bm{x}}}
\newcommand{\bfy}{{\bm{y}}}

\newcommand{\bfA}{\mathbf{A}}
\newcommand{\bfB}{\mathbf{B}}
\newcommand{\bfC}{\mathbf{C}}

\newcommand{\bfE}{\mathbf{E}}
\newcommand{\bfF}{\mathbf{F}}
\newcommand{\bfG}{\mathbf{G}}

\newcommand{\bfI}{\mathbf{I}}

\newcommand{\bfL}{\mathbf{L}}

\newcommand{\bfP}{\mathbf{P}}

\newcommand{\bfR}{\mathbf{R}}

\newcommand{\bfX}{\mathbf{X}}

\newcommand{\calC}{{\cal C}}
\newcommand{\calD}{{\cal D}}
\newcommand{\calE}{{\cal E}}

\newcommand{\calP}{{\cal P}}

\newcommand{\calS}{{\cal S}}

\newcommand{\calV}{{\cal V}}

\def\eg{\emph{e.g.}~}
\def\ie{\emph{i.e.}~}

\def\LGd{\dot{\bfL}_G}
\def\LG{\bfL_G}
\def\pdd{\ddot{\bfp}}

\def\qd{\dot{\q}}

\def\q{\bfq}

\newcommand{\bfPhi}{\boldsymbol{\Phi}}
\newcommand{\bfPsi}{\boldsymbol{\Psi}}

\newcommand{\defeq}{\stackrel{\mathrm{def}}{=}}

\usepackage[normalem]{ulem}

\newcommand{\ins}[1]{#1}
\newcommand{\del}[1]{}
\newcommand{\moved}[1]{}

\begin{document}

\title{ZMP support areas for multi-contact mobility \newline under frictional constraints}

\author{St\'{e}phane~Caron,~Quang-Cuong~Pham,~Yoshihiko Nakamura
\thanks{%
    St\'{e}phane Caron is with the Laboratoire d'Informatique, de Robotique et
    de Micro\'{e}lectronique de Montpellier (LIRMM), CNRS / Universit\'{e} de
    Montpellier.
    Quang-Cuong Pham is with the School of Mechanical and Aerospace
    Engineering, Nanyang Technological University, Singapore.
    Yoshihiko Nakamura is with the Department of
    Mechano-Informatics, University of Tokyo, Japan.

    Corresponding author:~\texttt{stephane.caron@normalesup.org}.}}

\maketitle

\begin{abstract}
    \del{In this paper,} We propose a method for checking and enforcing
    multi-contact stability based on the \ins{Zero-tilting Moment Point}
    (ZMP). The key to our development is the generalization of ZMP
    \emph{support areas} to take into account (a) frictional constraints and
    (b) multiple non-coplanar contacts. We introduce and investigate two kinds
    of ZMP support areas. First, we characterize and provide a fast geometric
    construction for the support area generated by valid contact forces, with
    no other constraint on the robot motion. We call this set the \emph{full
    support area}. Next, we consider the control of humanoid robots using the
    Linear Pendulum Mode (LPM). We observe that the constraints stemming from
    the LPM induce a shrinking of the support area, even \ins{for walking
    on horizontal floors} \del{which was previously unnoticed}. We propose an
    algorithm to compute the new area, which we call \emph{pendular support
    area}. We show that, \del{under LP control}\ins{in the LPM}, having the ZMP
    in the pendular support area is a necessary \emph{and sufficient} condition
    for contact stability. Based on these developments, we implement a
    whole-body controller and generate feasible multi-contact motions where an
    HRP-4 humanoid locomotes in challenging multi-contact scenarios.
\end{abstract}


\IEEEpeerreviewmaketitle

\section{Introduction}
\label{introduction}

The \ins{Zero-tilting Moment Point} (ZMP) is the dynamic quantity thanks to
which roboticists solved the problem of walking on horizontal floors. One of
its key properties is that \emph{dynamic stability}, \ie the balance of gravity
and inertial forces by \ins{valid} contact forces, implies that the ZMP lies in
the convex hull of ground contact points, the so-called \emph{support
area}~\cite{vukobratovic1972mb, goswami1999ijrr}. The support area thus
provides a necessary (non-sufficient) condition for contact stability on
horizontal floors.

For locomotion, the second key property of the ZMP lies in its coupling with
the position of the center of mass (COM). By keeping a constant angular
momentum and constraining the COM to lie on a plane, this relation simplifies
into the Linear Inverted Pendulum Mode (LIPM)~\ins{\cite{kajita1991icra,
kajita2001iros}}. In the LIPM, the COM moves away from the ZMP under the linear
dynamics of a point-mass at the tip of an inverted pendulum. The stabilization
problem is then to control the position of the tip (COM) of the pendulum by
moving its fulcrum (ZMP).

\begin{figure}[t]
    \centering
    \includegraphics[trim={0 0 0 0},clip,width=0.98 \columnwidth]{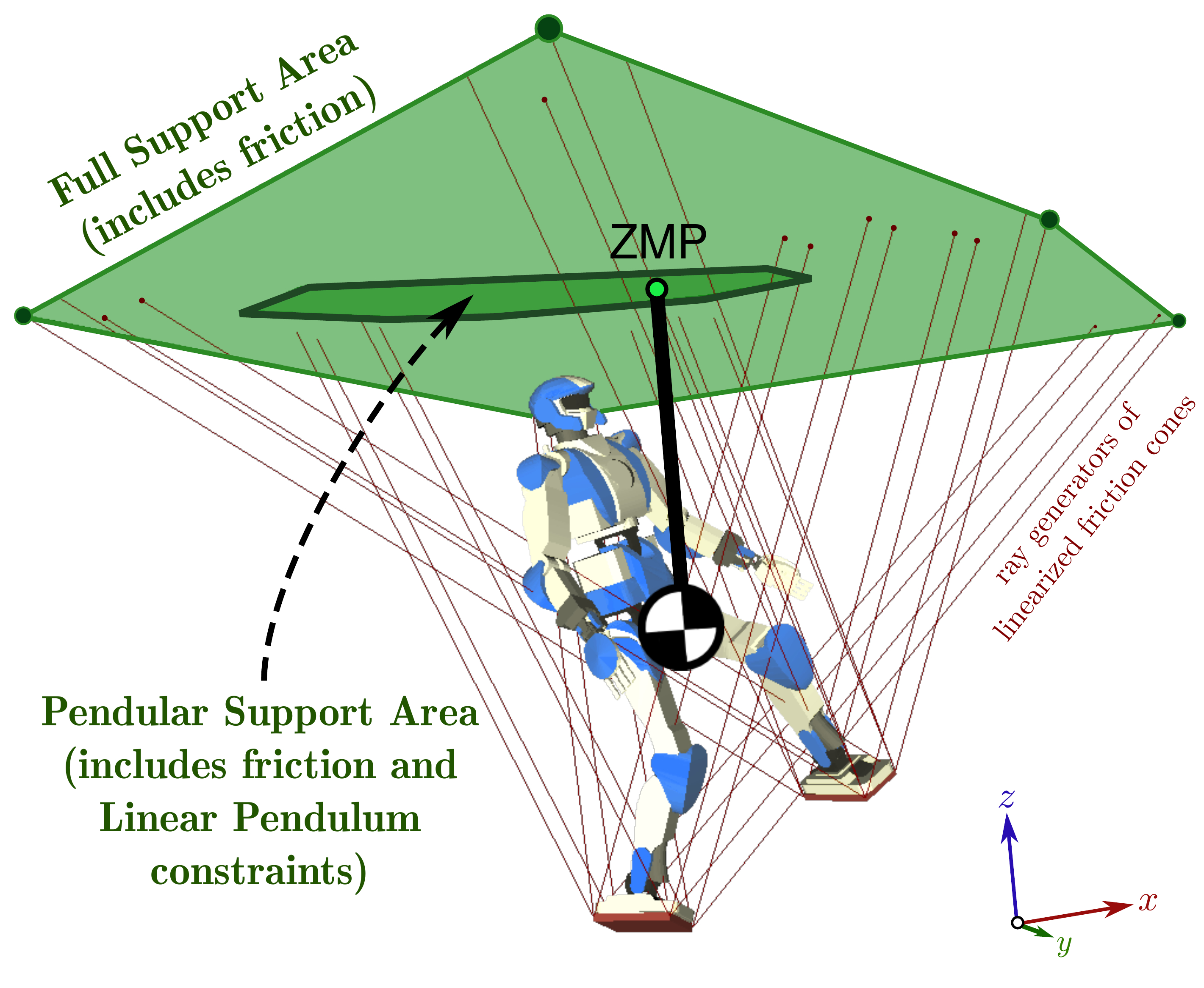}
    \caption{
        Overview of the construction proposed in this paper. The full ZMP
        support area, including \del{positive-}pressure and frictional constraints,
        is computed in an arbitrary virtual plane (here, above the robot's
        head). For locomotion, linearized pendulum dynamics are obtained by
        regulation of the angular momentum. The shrinking of the support area
        incurred by the Linear Pendulum Mode is fully taken into account. A
        whole-body controller based on these developments \del{finally }enables
        multi-contact locomotion \ins{in} arbitrary environments.
    }
    \label{fig:logo}
\end{figure}

These two main merits, a geometric stability condition and linearized
dynamics, are as well-known as the two main limitations of the classical ZMP:
it does not account for friction, and it can only be applied \ins{when all
contacts are coplanar}. The latter results from the definition of the ZMP as
the point \emph{on the floor}. In a general multi-contact scenario, each
contact defines its own surface and there is no single ``floor'' plane. In a
classic survey paper~\cite{sardain2004tsmc}, Sardain and Bessonnet stated the
problem to address as follows:

\begin{quote}
    \emph{The generalization of the ZMP concept [to the case of multiple
    non-coplanar contacts] would be actually complete if we could define what
    is the pseudo-support-polygon, a certain projection of the
    three-dimensional (3-D) convex hull (built from the two real support areas)
    onto the virtual surface, inside which the pseudo-ZMP stays.}
\end{quote}

In this paper, we construct the areas conjectured by Sardain and Bessonnet. Our
first contribution is to characterize the ZMP support area generated by valid
contact forces, with no other constraint on the robot motion. We call this set
the \emph{full support area}. Our analysis provides a geometric construction
that allows for fast calculations. Also, contrary to the assumption that
``friction limits are not violated'' usually made in the literature, this area
fully takes friction into account.

From a control point of view, locomoting systems usually regulate their linear
and angular momenta, as examplified by the Linear Pendulum Mode\del{l} where
the robot maintains a constant COM height and moves with a constant angular
momentum. These tasks \del{restrain}\ins{limit} the whole-body momentum of the
system, which consequently shrinks the ZMP support area. The second
contribution of this paper is an algorithm to compute the ZMP support area
\del{under linear-pendulum control}\ins{for the linear-pendulum mode}, which we
call \emph{pendular support area}. The latter takes into account both friction
and angular-momentum constraints on the robot motion\del{s}.

Combining these two advances, we design a whole-body controller for humanoids
locomoting on arbitrary terrains. Choosing a virtual plane above the COM, we
regulate the robot dynamics around that of a linear non-inverted pendulum. We
showcase the applicability of the controller by locomoting a model of the HRP-4
humanoid robot in a challenging multi-contact scenario involving a large step
supported by a hand-on-wall contact.

The paper is organized as follows. In Section~\ref{sec:background}, we review
previous work and introduce background notions necessary to understand the
present work. In Section~\ref{sec:contact}, we characterize the full support
area. We then provide in Section~\ref{sec:pendulum} an algorithm to calculate
the pendular support area, which we apply to multi-contact locomotion in
Section~\ref{sec:experiments}. Concluding remarks are finally provided in
Section~\ref{sec:conclusion}.

\section{Background}
\label{sec:background}

\subsection{Previous work}

\subsubsection{Stability criteria} on horizontal floors, the ZMP of dynamically
stable motions lies within the convex hull of contact points (CHCP). However,
when the robot makes contact with different non-coplanar surfaces, the ZMP can
no longer be defined as a point on the ``ground'' and the CHCP has no
established connection with dynamic stability. Various attempts have been made
in the literature to overcome this difficulty.

One line of research~\cite{sardain2004tsmc, mitobe2004iros, harada2006tro,
inomata2010iwamc} conjectured that the CHCP (a 3D volume in general) conveys
the stability condition, and consequently sought to define a new point lying
within this volume. Both~\cite{mitobe2004iros} and~\cite{inomata2010iwamc}
assumed that the moments at centers of pressure and ZMP are all zeros, which is
not the case in general\footnote{
    Sardain and Bessonnet~\cite{sardain2004tsmc} pointed out that the term
    ``zero moment point'' is misleading, as the moment at these points has
    actually a non-zero component along the surface normal. They suggested that
    ``zero-tilting moment point'' would be a more proper name.
} and thus results in a point that may not exist even in situations where
stability is possible. Harada et al.~\cite{harada2006tro} considered the CHCP
as a ZMP support volume when the robot makes two feet contact with a horizontal
floor and hand contacts with the environment. They detailed how to project the
support volume on the floor to obtain a ZMP support area. While their
construction applies to the general case with non-zero angular momentum, like
all approaches based on convex hull of contact points, it assumes infinite
friction coefficients. In this paper, we will see how to construct support
areas that also take friction into account.

A parallel line of research kept the ZMP in a plane but relaxed the constraint
that it coincides with the floor~\cite{kagami2002ar, sugihara2002icra,
shibuya2006iecon, sato2011tie}. Kagami et al.~\cite{kagami2002ar} had the
insight that the ZMP could be taken relative to any plane normal, yet still
assuming that this plane should pass through all contact points, which
restricted their scope to a maximum of three contacts points. Sugihara et
al.~\cite{sugihara2002icra} introduced the notion of ``Virtual Horizontal
Plane'' (VHP) in which contact points are projected on a virtual plane via the
line connecting them to the COM, and the convex hull of these points is then
taken as support area. Shibuya et al.~\cite{shibuya2006iecon} took the idea
further by considering a virtual plane \emph{above} the COM, while Sato et
al.~\cite{sato2011tie} applied this idea to stair climbing. However, like CHCP,
VHP support areas suppose infinite friction coefficients. The approach proposed
in the present paper also uses a virtual plane and the linear pendulum mode,
but our support areas fully account for friction, which makes them a necessary
\emph{and sufficient} condition for contact stability.

Breaking away from the notion of ZMP, another line of work has focused on
building criteria that keep equivalence with full contact
stability~\cite{saida2003icra, hirukawa2006icra, qiu2011isdhm, escande2013ras,
caron2015rss}. The seminal work by Saida et al.~\cite{saida2003icra} impulsed a
shift of paradigm from the ZMP to the gravito-inertial wrench. It also proposed
to orient the virtual plane orthogonally to the resultant force, which
reinstates the support area as convex hull of (projected) contact points -- an
idea that may have been overlooked by the literature so far. Next, Hirukawa et
al.~\cite{hirukawa2006icra} constructed the first full stability criterion for
the gravito-inertial wrench, yet with a high number of variables including all
contact forces. Later developments~\cite{qiu2011isdhm, escande2013ras,
caron2015rss} reduced these redundant variables to the gravito-inertial wrench
using the double-description method. Compared to traditional ZMP solutions,
this approach has the benefit of providing a full stability criterion, but at
the cost of the non-linear dynamics of the gravito-inertial wrench.
Furthermore, the nice geometric construction of the ZMP area is replaced by a
considerably less intuitive six-dimensional cone. The method that we introduce
in this paper reconciles full stability, linearized dynamics and a geometric
support area.

\subsubsection{Control} when it comes to control, the use of the ZMP is
historically tied to the LIPM, which was introduced in~\cite{kajita2001iros,
sugihara2002icra} and used in a wealth of subsequent
works~\cite{mitobe2004iros, harada2006tro, shibuya2006iecon, sato2011tie, kajita2003icra, harada2006ijhr,
morisawa2014sii, tedrake2015humanoids}. Kajita et
al.~\cite{kajita2003icra} brought in the technique of model predictive control
as a way to generate COM trajectories from desired ZMP positions. Harada et
al.~\cite{harada2006ijhr} proposed an analytical alternative with polynomial
solutions for the coupled COM-ZMP trajectories. Recently, Tedrake et
al.~\cite{tedrake2015humanoids} exhibited a closed-form solution for the
linear-quadratic regulator tracking a reference ZMP. However, all of these
methods only apply to locomotion on horizontal floors.

Aiming for locomotion on rough terrains, Zhao et al.~\cite{zhao2012humanoids}
extended the LIPM to a ``Prismatic Inverted Pendulum'' where the COM altitude
is allowed to vary linearly, while Morisawa et al.~\cite{morisawa2005icra}
provided a wider derivation where the COM belongs to a general two-dimensional
manifold. Rather than the ZMP, recent papers~\cite{morisawa2014sii,
englsberger2015tro} chose to control the Capture Point to stabilize the
unstable dynamics of the LIPM. In terms of support areas, though, the question
is the same for the Capture Point and the ZMP and was not addressed by these
developments. With the method proposed in the present paper, we realize a
control system with marginally stable dynamics using the ZMP as control point
and a linear pendulum mode. The benefit of our approach compared to these
previous works is that we are able to derive at the same time the support area
corresponding to our control variable.

As with stability criteria, solutions breaking away from control points were
also explored in the whole-body control literature. The main alternative is to
regulate contact forces directly, resulting in force distribution schemes where
desired contact forces and torques are tracked by a whole-body
controller~\cite{hirukawa2006icra, hyon2007tro, lee2010iros, ott2011humanoids,
righetti2013ijrr}. Force objectives can express whole-body tasks, such as
tracking of desired COM or angular momentum, as well as local ones, such as
minimizing friction forces~\cite{ott2011humanoids} or end-effector
torques~\cite{lee2010iros}. Notably, Righetti et al.~\cite{righetti2013ijrr}
characterized the class of force-distribution controllers for linear-quadratic
objectives in the absence of inequality constraints. Overall, force
distribution schemes yield fast computations and can cope with arbitrary
contact conditions, but they lack the foresight and intuition of methods based
on control points and support areas. Indeed, for locomotion, support areas
provide both reachable COM locations and a stability margin (the
point-to-boundary distance). Finding such indicators in the high-dimensional
contact-force space is still elusive. In recent developments,
\cite{nagasaka2012rs, audren2014iros} added a level of foresight to their
contact-force controllers via model-predictive control, while Zheng et
al.~\cite{zheng2015tro} constructed a metric that can be used as wrench-space
stability margin. In the present paper, we show that support areas can be
derived in arbitrary multi-contact configurations as well, providing both COM
reachability and stability margins suitable for locomotion.

\subsection{\ins{Newton-Euler} equations}

Let $m$ and $G$ represent the total mass and center-of-mass (COM) of the robot,
respectively. We write $\bfp_A$ the vector of absolute coordinates of a point
$A$ and denote by $O$ the origin of the absolute frame (so that $\bfp_O =
\bm{0}$). For a link~$k$, define:
\begin{itemize}
    \item $m_k$ the total mass of the link;
    \item $\bfp_{G_k}$ the vector of absolute coordinates of its COM
        $G_k$;
    \item $\bfR_k$ its orientation matrix in the absolute frame;
    \item $\bfomega_k$ its angular velocity in the link frame;
    \item $\bfI_k$ its inertia matrix in the link frame.
\end{itemize}
The linear momentum $\bfP$ and angular momentum $\LG$ of the robot, taken at
the COM $G$, are defined by:
\begin{eqnarray}
    \bfP &\defeq& \sum_{\mathrm{link}\,k} m_k \dot{\bfp}_{G_k},\\
    \LG &\defeq& \sum_{\mathrm{link}\,k} m_k\overrightarrow{GG_k}\times
    \dot{\bfp}_{G_k} + \bfR_k\bfI_k\bfomega_k.
\end{eqnarray}
\del{The \emph{dynamic wrench} of the robot at $G$ is the wrench $(\dot{\bfP},
\LGd)$ obtained by differentiation over time of the robot's momentum $(\bfP,
\bfL_G)$. It is a purely kinematic object that can be computed by forward
kinematics from joint-angle positions, velocities and accelerations.}

The fundamental principle of dynamics states that the \ins{rate of change
of the momentum} is equal to the total wrench of forces acting on the system,
that is:
\begin{equation}
    \label{newton-euler}
    \left[ \begin{array}{c} 
            \dot{\bfP} \\ 
            \LGd 
    \end{array} \right]
    \ = \
    \left[ \begin{array}{c} 
            \bff^g \\ 
            {\bm{0}} 
    \end{array}\right]
    \, + \, 
    \sum_{\mathrm{contact}\, i}
    \left[\begin{array}{c} 
            \bff_{i} \\ 
            {\overrightarrow{GC}_{i}} \times \bff_{i}
    \end{array}\right],
\end{equation}
where $\bff^g$ denotes the gravity force, $C_i$ the $i^\textrm{th}$ contact
point and $\bff_{i}$ the contact force exerted by the environment on the robot
at $C_i$. \ins{Equation~\eqref{newton-euler} is called the Newton-Euler
equations} of the system, sometimes also referred to as ``dynamic balance'' or
the ``dynamic equilibrium''. It can be equivalently derived from Gauss's
principle of least constraint, and corresponds to the six unactuated components
in the equations of motion of the system $\{$robot~$+$~environment$\}$
\cite{wieber2006}. 

Define the \emph{gravito-inertial wrench}, taken this time at $O$:
\begin{equation}
    \label{def-gi}
    \bfw^{gi}_O
    \ \defeq \
    \left[ \begin{array}{c} 
        \bff^{gi} \\ 
        \bftau^{gi}_O
    \end{array} \right]
    \ \defeq \ 
    \left[\begin{array}{c} 
            \bff^g - \dot{\bfP} \\ 
            \bfp_G \times (\bff^g - \dot{\bfP}) - \LGd
    \end{array}\right].
\end{equation}
Define similarly the \emph{contact wrench}:
\begin{equation}
    \label{def-c}
    \bfw^{c}_O
    \ \defeq \
    \left[ \begin{array}{c} 
        \bff^{c} \\ 
        \bftau^{c}_O
    \end{array} \right]
    \ \defeq \ 
    \sum_{\mathrm{contact}\, i}
    \left[\begin{array}{c} 
            \bff_{i} \\ 
            \bfp_{C_i} \times \bff_{i}
    \end{array}\right]
\end{equation}
The Newton-Euler equation~\eqref{newton-euler} can be written in terms of these two
wrenches as:
\begin{equation}
    \label{dyneq}
    \bfw^{gi} + \bfw^c = \bm0.
\end{equation}
\del{This formulation separates whole-body accelerations ($\bfw^{gi}$),
over which the robot has direct control, from interaction forces with the
environment ($\bfw^c$), which cannot be controlled directly. The former are
commonly measured by inertial measurement units and the latter by force
sensors.} \ins{This formulation separates spatial accelerations
($\bfw^{gi}$), that are usually measured by inertial measurement units, from
interaction forces with the environment ($\bfw^c$), usually measured by force
sensors. Since the two wrenches are simply opposites, we will only use the
contact wrench $(\bff, \bftau_O)$ in the following calculations, dropping the
superscript $c$ to alleviate notations.}

\subsection{Contact stability}

We assume that all contacts between the environment and the robot are
\emph{surface} contacts\del{ between polyhedral rigid bodies}\ins{, \ie contacts
between a flat surface of the robot and a flat surface of the environment}.
\ins{Wrenches exerted on each contacting link} are then fully described by
applying contact forces at the vertices of the \del{link's }contact
polygon~\cite{caron2015icra}, which warrants the formulation by contact points
that we have followed so far. Define the \emph{contact normal} $\bfn_i$ at
$C_i$ as the normal to the contact surface pointing from the environment
towards the contacting link. Under Coulomb's model of dry friction, contact
forces $\bff_i$ lie inside a friction cone directed by $\bfn_i$:
\begin{eqnarray}
    \label{coulomb-l2}
    \left\| \bfn_i \times \bff^{c}_i \times \bfn_i \right\|_2 & \leq & \mu_i (\bff^{c}_i \cdot \bfn_i).
\end{eqnarray}
Frictional constraints restrict the range of contact wrenches $\bfw^c$ that the
robot can generate without breaking any contact:~when each contact force lies
in a cone $\calC_i$, the contact wrench lies in the Contact Wrench Cone (CWC)
$\calC^c$ projected from all $\calC_i$'s via the mapping \eqref{def-c}. Because
of the connection~\eqref{dyneq}, the gravito-inertial wrench belongs to
Gravito-inertial Wrench Cone (GIWC) $\calC^{gi} = -\calC^c$. This link between
\ins{feasible} motions and \ins{keeping} contacts is embodied by the
notion of contact stability:

\begin{definition}[Weak contact stability~\cite{pang2000zamm,balkcom2002ijrr}]
    A motion of the robot is \emph{(weak-contact) stable} if and only if the
    contact wrench it generates belongs to the CWC.
\end{definition}
\del{We will equivalently say that a motion is \emph{dynamically stable}
when the gravito-inertial wrench it generates belongs to the GIWC. Observe
that, since the gravito-inertial and contact wrenches are simply
opposites~\eqref{dyneq}, the two expressions refer to the same concept. From
now on, we will always use the contact wrench $(\bff, \bftau_O)$, dropping the
superscript $c$ to alleviate notations.} Weak contact stability is the
underlying stability criterion used in recent multi-contact
developments~\cite{hirukawa2006icra, qiu2011isdhm, escande2013ras,
caron2015rss}. Static stability~\cite{bretl2008tro} corresponds to contact
stability when the \del{dynamic wrench}\ins{whole-body momentum} is zero. Note
that the term ``stability'' is used here in the sense defined by Pang and
Trinkle~\cite{pang2000zamm}. It should not be confused with the (a-priori
unrelated) notion of Lyapunov stability.

\subsection{Linearized wrench cones}

We linearize the quadratic constraint~\eqref{coulomb-l2} by approximating
friction cones by friction pyramids \del{(also known as polyhedral convex
cones}. Denoting by $\ins{\bff_{ij}}$ the ray vectors of the latter, the constraint
becomes:
\begin{equation}
    \label{coulomb-l1}
    \bff_{i} = \sum_{\mathrm{ray}\,j} \lambda_{ij} \, \ins{\bff_{ij}}, \quad \lambda_{ij} \geq 0.
\end{equation}
The set of ray vectors $\{ \ins{\bff_{ij}} \}$ is known as the \emph{span}
representation of the pyramidal cone. It can be computed directly from the
contact frame and friction coefficient $\mu_i$. For example, the expression of
a four-sided pyramid is $\left\{\bfn_i \pm \frac{\mu_i}{\sqrt{2}} \, \bft_i \pm
\frac{\mu_i}{\sqrt{2}} \, \bfb_i\right\}$, with $(\bft_i, \bfb_i, \bfn_i)$ an
orthonormal contact frame, \ins{where the normalization by $\sqrt{2}$ is added to
make the friction pyramid an inner approximation of the friction cone}.
Injecting~\eqref{coulomb-l1} into Equation~\eqref{def-c} yields a span
representation for the contact wrench cone:
\begin{equation}
    \left[ \begin{array}{c} 
        \bff \\ 
        \bftau_O
    \end{array} \right]
    \ = \
    \sum_{i, j}
    \lambda_{ij}
    \left[\begin{array}{c} 
            \ins{\bff_{ij}} \\ 
            \bfp_{C_i} \times \ins{\bff_{ij}}
    \end{array}\right]
    \quad \lambda_{ij} \geq 0.
\end{equation}
Let us define $\bftau_{O,ij} = \bfp_{C_i} \times \ins{\bfu_{ij}}$. After re-indexing
the couples $i,j$ into a single index $i$ (counting the same contact point $C_i$
multiple times accordingly), we get:
\begin{equation}
    \label{eq:span-c}
    \left[ \begin{array}{c} 
        \bff \\ 
        \bftau_O
    \end{array} \right]
    \ = \
    \sum_i \lambda_i \left[\begin{array}{c} \bff_i \\ \bftau_{O, i} \end{array} \right]
    , \quad \lambda_i \geq 0.
\end{equation}
We have thus obtained the span representation of the CWC. A motion is
weak-contact stable if and only if its contact wrench can be written
as~\eqref{eq:span-c} for a certain set of coefficients $\lambda_i \geq 0$.

\section{Full ZMP support area}
\label{sec:contact}

Let $\bfn$ be a fixed unit space vector, not necessarily aligned with the
gravity vector. The \ins{Zero-tilting Moment Point} (ZMP) is a point $Z$
where the moment of the contact \del{(equivalently, gravito-inertial)}
wrench aligns with $\bfn$~\cite{sardain2004tsmc}, that is, $\bfn \times
\bftau_Z = \bm0$. Consequently,
\begin{eqnarray}
    -\bfn \times (\bfp_Z \times \bff) + \bfn \times \bftau_O & = & \bm0 \\
    -(\bfn \cdot \bff) \bfp_Z + (\bfn \cdot \bfp_Z)
    \bff + \bfn \times \bftau_O  & = & \bm0
\end{eqnarray}
We are interested in computing the ZMP in the plane that contains $O$ and that
is orthogonal to $\bfn$, hereafter denoted by $\Pi(O,\bfn)$. The relation $Z
\in \Pi(O, \bfn)$ is expressed by $\bfn \cdot \ins{\bfp_Z}=0$, so that
the equation above becomes:
\begin{equation}
    \label{eq:zmp}
    \bfp_Z \ = \ \frac{\bfn \times \bftau_O}{\bfn \cdot \bff}
\end{equation}
On horizontal floors, $O$ is taken on the floor and $\bfn$ is upward vertical,
so that $\Pi(O, \bfn)$ coincides with the floor plane. However, in what follows
we assume that $O$ can be located at an arbitrary fixed position in space,
while $\bfn$ can be an arbitrary unit vector.

\moved{Note how, when $\bfn \cdot \bff = 0$, Equation~\eqref{eq:zmp} has no solution
and cannot be used to define a point $Z$. This singularity is present on
horizontal floors as well, where the division by zero occurs when $\bff$ is
horizontal.}

\subsection{Construction of the full support area}
\label{sec:support-area}

Equation~\eqref{eq:zmp} presents the ZMP as a two-dimensional projection of the
contact wrench $(\bff, \bftau_O)$. Since contact stability is characterized by
the CWC, we define the support area of the ZMP as the image of the CWC by
this projection:
\begin{definition}[Full support area]
    The \emph{full support area} $\calS$ of the ZMP in the plane $\Pi(O, \bfn)$ is
    the image of the CWC by the projection~\eqref{eq:zmp}:
    \begin{equation}
        \calS = \left\{\left. \bfp_Z = \frac{\bfn \times \bftau_O}{\bfn \cdot \bff} \
            \right| \ (\bff,
        \bftau_O) \in \calC^c \right\}
    \end{equation}
\end{definition}
The key idea to calculate this area is to use the span
representation~\eqref{eq:span-c} of the CWC, which enables rewriting
Equation~\eqref{eq:zmp} as
\begin{equation}
    \label{zmp-e2}
    \bfp_Z \ = \ \frac{\sum_i \lambda_i (\bfn \times \bftau_{O,i})}{\sum_i
      \lambda_i (\bfn \cdot \bff_i)}, \quad \lambda_i \geq 0.
\end{equation}
Next, define the points $Z_i$ by:
\begin{equation}
    \label{eq:Zi}
    \bfp_{Z_i} \ \defeq \ \frac{\bfn \times \bftau_{O,i}}{\bfn \cdot \bff_i}
\end{equation}
and denote by $p_i \defeq (\bfn \cdot \bff_i)$ the \emph{virtual pressure} of the
contact force generator $\bff_i$ through the virtual plane. Then,
\begin{equation}
    \label{eq:zcomb}
    \bfp_Z \ = \ \frac{\sum_i \lambda_i p_i \, \bfp_{Z_i}}{\sum_i \lambda_i p_i}, \quad \lambda_i \geq 0.
\end{equation}

On horizontal floors, $\bfn$ and all contact forces $\bff_i$ point upwards, so
that all pressures $p_i$ are positive. This makes the ZMP $Z$ a convex
combination of the $Z_i$'s from Equation~\eqref{eq:zcomb}. Furthermore,
Equation~\eqref{eq:Zi} simplifies to $Z_i = C_i$, \ie, the vertices of the
support area $\calS$ coincide with contact points. Our definition of the full
support area therefore coincides with the conventional support area on
horizontal floors.

In general, however, virtual pressures $p_i$ can be either positive or
negative.\footnote{We assume $\bfn$ is chosen so that none of them is zero,
which is easy to do since there is only a finite set of generators $\{ \bff_i
\}$.} Let us then partition the set of generator indices $I$ into $I^+ \defeq
\{ i \,|\, p_i > 0 \}$ and $I^- \defeq \{ i \,|\, p_i < 0\}$. For any $K
\subset I$, denote by $\sigma(K) \ \defeq \ \sum_{i \in K} \lambda_i |p_i|$ and
define
\begin{eqnarray}
    \alpha_i & \defeq & \frac{+\lambda_i p_i}{\sigma(I^+)} \ \textrm{ for } i \in I^+,
    \quad 
    \quad 
    \alpha \ \defeq \ \frac{\sigma(I^+)}{\sigma(I)}. \\
    \beta_i & \defeq & \frac{-\lambda_i p_i}{\sigma(I^-)} \ \textrm{ for } i \in I^-,
    \quad 
    \quad 
    \beta \ \defeq \ \frac{\sigma(I^-)}{\sigma(I)}.
\end{eqnarray}
Equation~\eqref{zmp-e2} becomes
\begin{equation}
    \label{eq:above}
    \bfp_Z \ = \ \frac{1}{\alpha - \beta} \left[\alpha \sum_{i \in I^+}
    \alpha_i \bfp_{Z_i} - \beta \sum_{i \in I^-} \beta_i \bfp_{Z_i}\right].
\end{equation}

Define the \emph{positive-pressure polygon} as the convex hull of $Z_i$'s for
$i \in I^+$: $\calP^+ \defeq \{\sum_{i \in I^+} {\alpha}_i \bfp_{Z_i}, \alpha_i
\geq 0, \sum_i {\alpha}_i = 1\}$, and define the negative-pressure polygon
$\calP^-$ similarly. Equation~\eqref{eq:above} can be rewritten as
\begin{equation}
    \label{zmp-e3}
    \bfp_Z \ = \ \frac{\alpha \bfp_{Z^+} - \beta \bfp_{Z^-}}{\alpha - \beta},
\end{equation}
where $\alpha \geq 0$, $\beta \geq 0$, $Z^+ \in \calP^+$ and $Z^- \in \calP^-$. 
Next, let
\begin{equation}
    \calP^+ - \calP^- = \left\{\left. \bfp_{Z^+} - \bfp_{Z^-} 
        \ \right| \ 
        Z^+ \in \calP^+, \ Z^- \in \calP^-\right\}
\end{equation}
denote the Minkowski difference of $\calP^+$ and $\calP^-$. It is a polygon as
difference of two polygons, so that it admits a span representation $\calP^+ -
\calP^- = \textsc{conv}(\{ \bfr_1, \ldots, \bfr_k \})$ as convex hull
($\textsc{conv}$) of a set of vertices. We can now characterize the support
area as follows:

\begin{proposition}
    \label{prop:double-cone}
    If all virtual pressures $p_i$ have the same sign, then the full support
    area $\calS$ is the convex hull of the vertices $\{Z_i\}$. Otherwise,
    when both $\calP^+$ and $\calP^-$ are nonempty, let
    \begin{equation}
        \calP^+ - \calP^- = \textsc{conv}(\{ \bfr_1, \ldots, \bfr_k \})
    \end{equation}
    denote
    their Minkowski difference. The full support area $\calS$ is then the
    reunion of two polygonal cones $\calC^+$ and $\calC^-$ given by
    \begin{eqnarray}
        \calC^+ & = & \calP^+ + \textstyle \sum_i \mathbb{R}^+ \bfr_i, \\
        \calC^- & = & \calP^- + \textstyle \sum_i \mathbb{R}^+ (-\bfr_i).
    \end{eqnarray}
    In particular, when $\calP^+$ and $\calP^-$ inter\del{e}sect with nonempty
    interior, the full support area $\calS$ spans the whole virtual plane
    $\Pi(O, \bfn)$.
\end{proposition}

\begin{figure}[t]
    \centering
    \includegraphics[height=5cm]{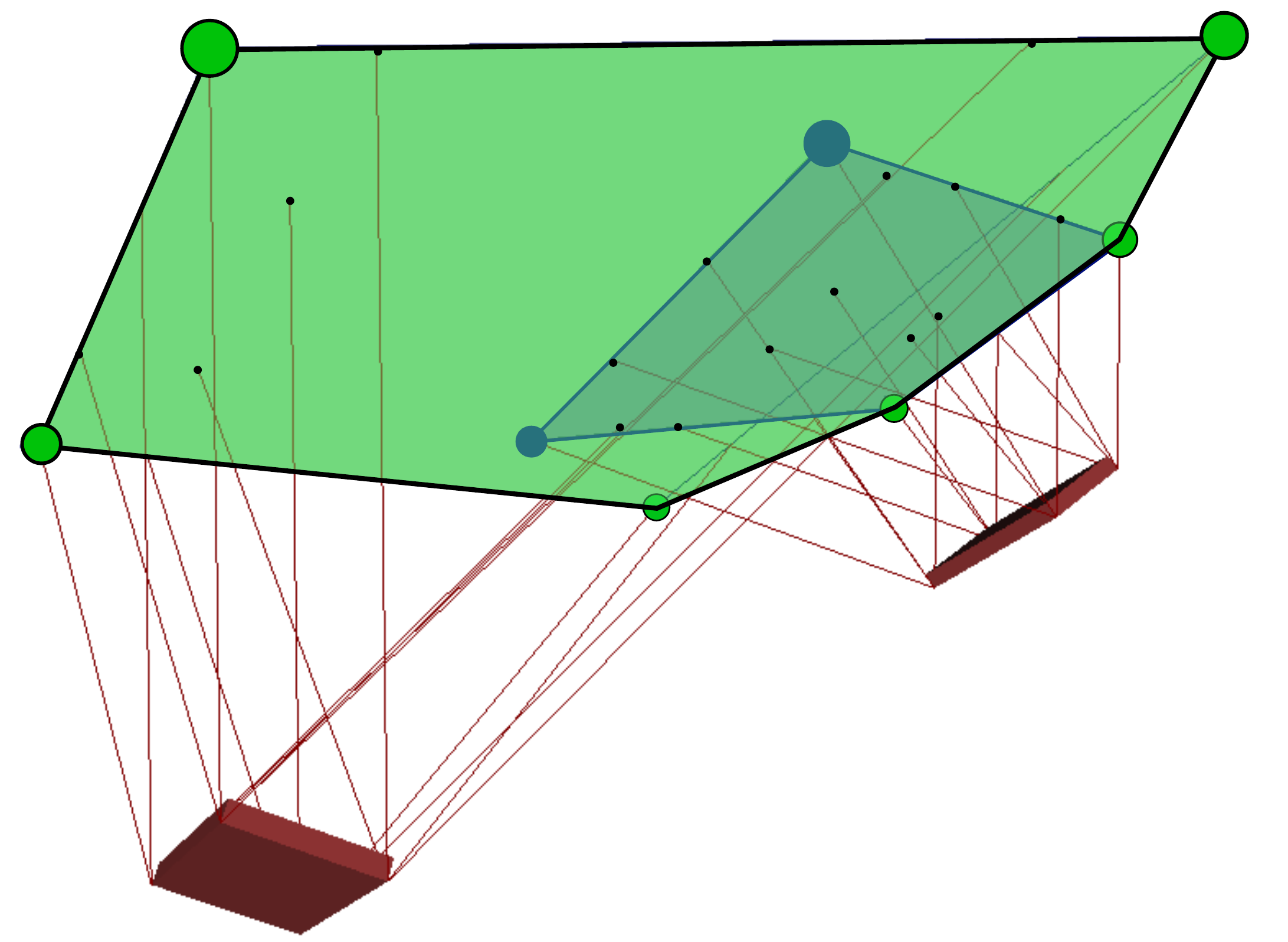}
    \caption{
        Geometric construction of the ZMP support area in the polygonal case.
        Ray generators of friction cones (red lines) are traced until they
        intersect the virtual plane, yielding points inside the support area
        (black dots). The support area (in green) is the convex hull of this
        set of points (Proposition~\ref{prop:inter-poly}). The blue polygon
        corresponds to the individual support polygon of the right contact.
        These polygons describe the contribution of each contact to the
        expansion of the support area, and can \eg be used in contact planning
        to evaluate how new contact candidates would expand the area.
    }
    \label{fig:poly-case}
\end{figure}

\begin{IEEEproof}
    \ins{(Note that this proof is quite formal and may be skipped by first-time
    readers.)} \ins{The main idea of the proof is to combine pairs of points
    in $\calP^+ \times \calP^-$ to generate rays in $\calC^+$ and $\calC^-$.
    Let us first clear out simple cases. When}\del{Suppose that} all $p_i$'s
    are positive ($\calP^- = \emptyset$)\del{. Then}, Equation~\eqref{eq:zcomb}
    shows that $\bfp_Z$ is a convex combination of the vertices of $\calP^+$,
    where the weights $(\lambda_i p_i) / (\sum_j \lambda_j p_j)$ can be chosen
    freely, so that $\calS = \calP^+$. The case where $\calP^+ = \emptyset$ is
    treated identically. Next, suppose that both $\calP^+$ and $\calP^-$ are
    nonempty. Equation~\eqref{zmp-e3} can be reformulated as
    \begin{equation}
        \bfp_Z 
        \: = \: \bfp_{Z^+} + \frac{\beta}{\alpha - \beta}\overrightarrow{Z^-Z^+} 
        \: = \: \bfp_{Z^-} + \frac{\alpha}{\beta - \alpha}\overrightarrow{Z^+Z^-}
    \end{equation}
    \ins{where $\alpha$ (resp. $\beta$) is the weight of $\calP^+$ (resp.
    $\calP^-$) in Equation~\eqref{zmp-e3}}. Therefore, the set of points $Z$
    defined by this equation is
    \begin{align}
        \calS & \defeq \nonumber
        \left\{\bfp_{Z^+} + \frac{\beta}{\alpha - \beta}\overrightarrow{Z^-Z^+}, 
        \ \alpha \geq \beta \geq 0, Z^\pm \in \calP^\pm \right\} \\
        & \cup
        \left\{\bfp_{Z^-} + \frac{\alpha}{\beta - \alpha}\overrightarrow{Z^+Z^-}, 
        \ \beta \geq \alpha \geq 0, Z^\pm \in \calP^\pm \right\}
    \end{align}
    Given the orderings of $\alpha$ and $\beta$, we can further simplify the ratios
    into a single positive scalar, so that $\calS = \calC^+ \cup \calC^-$ with
    \begin{eqnarray}
        \label{cone-plus}
        \calC^+ & = & 
        \left\{\bfp_{Z^+} + \lambda \overrightarrow{Z^-Z^+},
        \ \lambda \geq 0, Z^\pm \in \calP^\pm\right\},
        \\
        \calC^- & = &
        \left\{\bfp_{Z^-} + \lambda \overrightarrow{Z^+Z^-},
        \ \lambda \geq 0, Z^\pm \in \calP^\pm\right\}.
    \end{eqnarray}
    The set $\calD = \calP^+ - \calP^-$ is a convex polygon as Minkowski
    difference of two convex polygons.
    To conclude, we show that $\calC^+ = \calP^+ + \mathbb{R}^+ \calD$. The
    inclusion $\subset$ is straightforward
    from~\eqref{cone-plus}. Now, let 
    \begin{equation}
        \bfp_C = \bfp_{Z_0^+} + \mu (\bfp_{Z_1^+} - \bfp_{Z^-})
    \end{equation}
    denote any point in $\calP^+ + \mathbb{R}^+ \calD$. Define
    \begin{equation}
        \bfp_{Z^+} \ \defeq \ \frac{1}{1 + \mu} \, \bfp_{Z_0^+} + \frac{\mu}{1 + \mu} \, \bfp_{Z_1^+}.
    \end{equation}
    One can check that $\bfp_C = \bfp_{Z^+} + \mu(\bfp_{Z^+} - \bfp_{Z^-})$, where
    $Z^+$ belongs to $\calP^+$ as convex combination of two points from this convex
    polygon. Thus $C \in \calC^+$, which establishes the converse inclusion
    $\supset$.

    Finally, note how, when $\calP^+ \cap \calP^-$ has non-empty interior,
    $\calD$ contains a neighborhood of the origin. For any pair of points $(A,
    Z^+) \in \Pi(O, \bfn) \times \calP^+$, this implies that there exists a
    scaling $\epsilon > 0$ such that $\epsilon (\bfp_A - \bfp_{Z^+}) \in
    \calD$. Then, as $\calC^+ = \calP^+ + \mathbb{R}^+ \calD$, we have
    $\bfp_{Z^+} + \frac{1}{\epsilon} (\epsilon (\bfp_A - \bfp_{Z^+})) = \bfp_A
    \in \calC^+$. Since $A$ was taken arbitrarily in the virtual plane, this
    establishes that $\calS = \Pi(O, \bfn)$, \ie the support area spans the
    whole virtual plane.
\end{IEEEproof}

\begin{figure}[t]
    \centering
    \includegraphics[height=5cm]{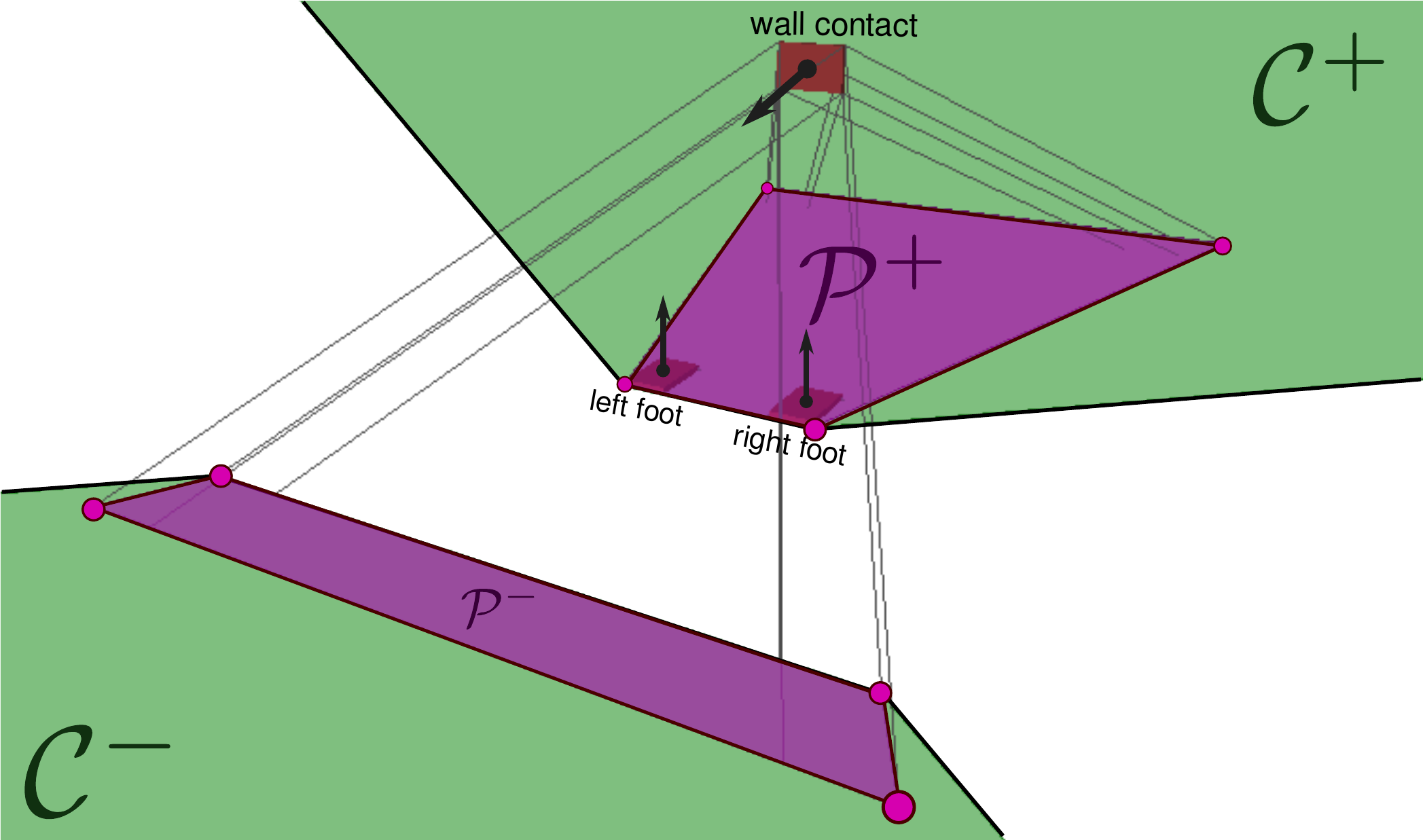}
    \caption{
        Example where the full support area $\calS$ is the union of two cones
        $\calC^+$ and $\calC^-$ (in green). There are three contacts\del{ in total}:
        two feet \del{contacts} on a horizontal floor, and a wall contact
        located 50 cm forward and 90 cm above ground. The virtual plane is
        taken in the feet's \del{horizontal }plane. Polygons $\calP^+$ and
        $\calP^-$ used in the geometric construction\ins{ of the support area}
        are drawn in purple, while the rays of friction cones are depicted by
        gray lines. In this example, $\calC^+$ \del{illustrates}\ins{shows} how
        the conventional support polygon between the two feet is extended by
        the wall contact, while $\calC^-$ \del{shows how}\ins{is a
        complementary support area resulting from} the ability \del{for the
        robot }to ``push \del{itself }down\ins{ on the wall}'' \del{(using the
        wall contact)}.
    }
    \label{fig:cone-case}
\end{figure}

The geometric construction given by Proposition~\ref{prop:double-cone} provides
fast computations \ins{of} the full support area. \del{On an average laptop
computer (Intel Core i7-3540M CPU @ 3.00 GHz), an unoptimized Python script
could compute it in real time at around 120 Hz. See the code accompanying this
paper [36].} In particular, it is faster to use this approach than to project
the CWC computed by polyhedral duality methods~\cite{caron2015rss}. (A
comparison of computation times for both approaches is reported in
Table~\ref{table:times}.)

\subsection{Geometric properties}

Proposition~\ref{prop:double-cone} gives the span representation of the full
support area. By construction, this area depends only on contact locations $\{
\bfp_{C_i} \}$ and on the choice of the virtual plane $\Pi(O, \bfn)$. The
latter involves the choice of a fixed point $O$, but we will now see that the
location of this point only matters as far as it defines the virtual plane
$\Pi(O, \bfn)$ of the ZMP.

\begin{proposition}
    \label{prop:invar2d}
    The support area $\calS$ does not depend on the coordinates of the
    reference point $O$ in the virtual plane $\Pi(O, \bfn)$.
\end{proposition}

\begin{IEEEproof}
By definition of the support area,
\begin{equation}
  \label{eq:Z}
  \textstyle
  \calS = \left\{Z \in \Pi(O,\bfn): \overrightarrow{OZ} = \frac{\bfn \times
      \bftau_O}{\bfn \cdot \bff} \right\}
\end{equation} 
where $(\bff, \bftau_O) \in \calC^c$ ranges over the CWC. Now, choose a point
$O'\in\Pi(O,\bfn)$ and consider
\begin{equation} 
    \label{eq:Z'} 
    \textstyle \calS' = \left\{Z' \in \Pi(O,\bfn) = \Pi(O',\bfn):
        \overrightarrow{O'Z'} = \frac{\bfn \times \bftau_{O'}}{\bfn \cdot \bff} \right\}.
\end{equation}
Given a contact wrench $(\bff, \bftau_O)$, we have
\begin{eqnarray}
    \overrightarrow{O'Z'} & = & \frac{\bfn \times \bftau_{O'}}{\bfn \cdot \bff}
    = \frac{\bfn \times (\overrightarrow{O'O} \times \bff) + \bfn \times \bftau_{O}}{\bfn \cdot \bff}\\ 
    \overrightarrow{O'Z'} & = & \overrightarrow{O'O} - (\bfn \cdot \overrightarrow{O'O})
    \frac{\bff}{\bfn \cdot \bff} + \overrightarrow{OZ}= \overrightarrow{O'Z}.
\end{eqnarray}
Thus, $Z'=Z$, and since the wrench we considered is arbitrary, we have shown
that $\calS=\calS'$.
\end{IEEEproof}

\begin{figure}[t]
    \centering
    \includegraphics[height=5cm]{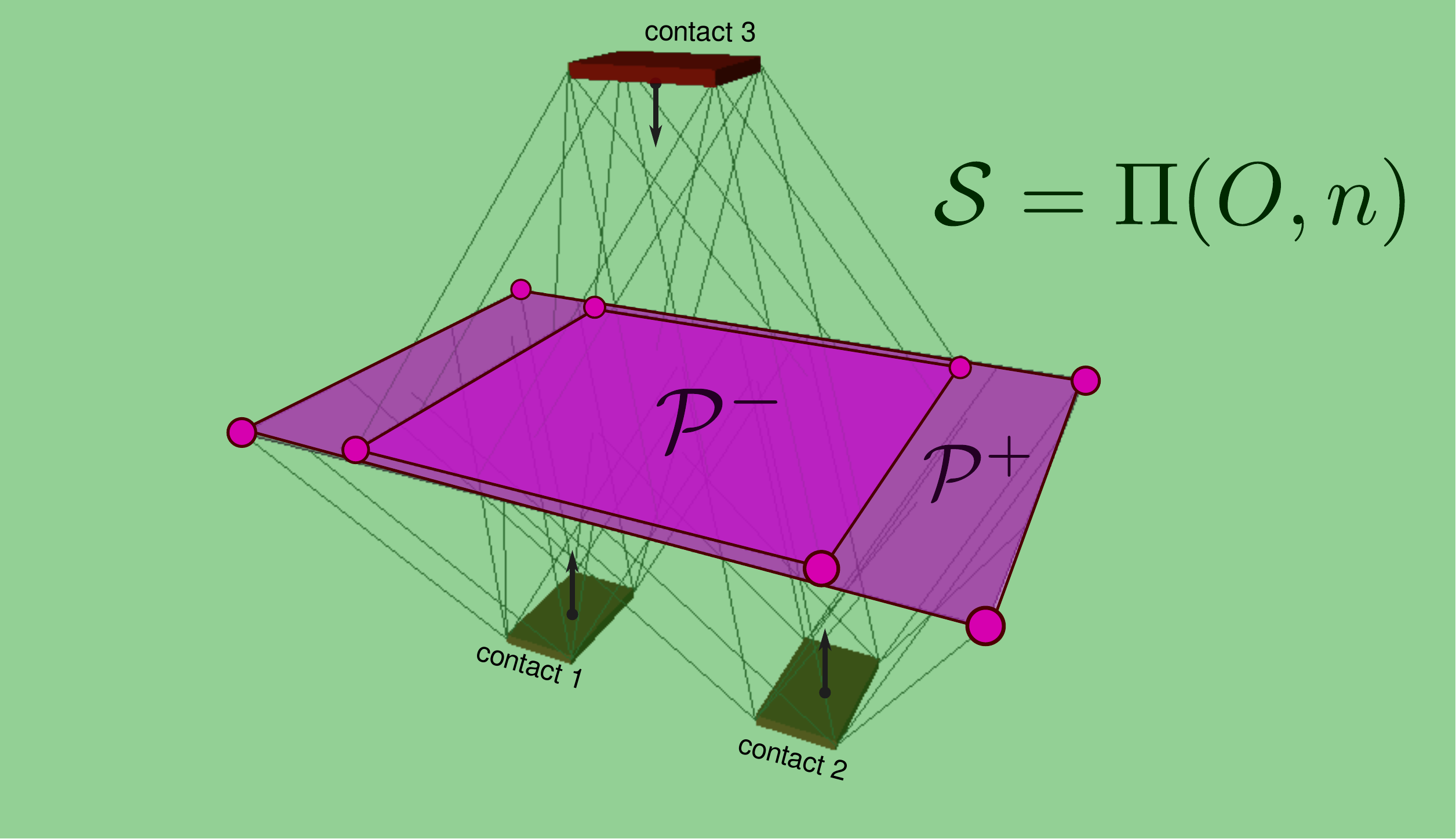}
    \caption{
        Example where the support area spans the whole virtual plane. There are
        three contacts in total. Contacts 1 and 2 have an upward vertical
        normal and correspond \eg to two feet on horizontal floor. Contact 3
        is located one meter above the others and has a \emph{downward}
        vertical normal, corresponding \eg to a hand pushing on the ceiling
        above the robot. The virtual plane is taken 50 cm above contacts 1 and
        2 and 50 cm below contact 3. Gray lines represent the ray generators of
        friction cones. In \del{such a}\ins{this} setting, the robot's contacts
        are in force closure: they can generate any resultant \del{contact
        }wrench, hence any ZMP.
    }
    \label{fig:full-case}
\end{figure}

By contrast, the support area does change for displacements of $O$ along the
plane normal $\bfn$. Let us analyze the impact of this remaining coordinate by
relaxing the assumption that $(\bfn \cdot \bfp_Z) = 0$. We now denote by $d_Z$
the coordinate of the virtual plane, so that $d_Z \defeq (\bfn \cdot \bfp_Z)$ and
$d_G \defeq (\bfn \cdot \bfp_G)$. On a horizontal floor, $d_Z$ and $d_G$
correspond to the altitude of the ZMP and COM, respectively. Given
Proposition~\ref{prop:invar2d}, we can write the virtual plane $\Pi(d_Z,
\bfn)$. The definition $\bfn \times \bftau_Z = \bm{0}$ of the ZMP yields:
\begin{equation}
    \label{eq:zmp-dZ}
    \bfp_Z \ = \ \frac{\bfn \times \bftau_O}{\bfn \cdot \bff} + d_Z \frac{\bff}{\bfn \cdot \bff}.
\end{equation}
And repeating the step from Equation~\eqref{eq:Zi},
\begin{eqnarray}
    \bfp_{Z_i} & \defeq & \frac{\bfn \times (\bfp_{C_i} \times \bff_i)}{\bfn \cdot \bff_i} + d_Z \frac{\bff_i}{\bfn \cdot \bff_i}, \\
    \bfp_{Z_i} & = & \bfp_{C_i} + (d_Z - d_i) \frac{\bff_i}{\bfn \cdot \bff_i}. \label{zi-dz}
\end{eqnarray}
We see from this equation that the vertices $Z_i$ are located at the
intersection between the plane $\Pi(d_Z, \bfn)$ and the ray $(C_i, \bff_i)$ of
the linearized friction cone, which establishes that:

\begin{proposition}
    \label{prop:vertices}
    The vertices of the support area are located at the intersection between
    the virtual plane and the rays of the friction cones.
\end{proposition}

\del{This proposition gives a first geometric interpretation of the support area,
yet it only provides its vertices. We saw in Proposition~\ref{prop:double-cone}
how reconstructing the area from these vertices is not straightforward. Still,
when a suitable plane normal $\bfn$ can be found so that $\calS$ is a polygon,
we get a simple geometric characterization:}
\ins{Combining Propositions~\ref{prop:double-cone} and \ref{prop:vertices}, we
get:}

\begin{proposition}[Polygonal case]
    \label{prop:inter-poly}
    When all virtual pressures $p_i = (\bfn \cdot \bff_i)$ \del{are
    positive} \ins{have the same sign}, the support area is the convex hull
    of the intersection\ins{s} between linearized friction cones and the virtual
    plane.
\end{proposition}

This result is coherent with the horizontal-floor setting where the virtual
plane intersects friction cones at their apexes (\ie at contact points) and
virtual pressures are all positive from contact unilaterality. A condition
similar to the positivity of virtual pressures was also observed by Saida et
al. (Proposition 4.1 in~\cite{saida2003icra}) for some plane components of the
contact wrench. 

Figure~\ref{fig:poly-case} illustrates the geometric construction in the
polygonal case. The support area (in green) is the convex hull of black points
projected from friction rays (red lines). Alternatively, each individual
contact surface projects its own support polygon (blue polygon in
Figure~\ref{fig:poly-case}), and the ZMP support area is the convex hull of
these individual polygons. This second construction can be useful for contact
planning, where the quality of contact candidates can be assessed by the
expansion that they bring to the support area.

Figure~\ref{fig:cone-case} shows a configuration where the support area is the
union of two polygonal cones. In this setting, the robot has its two feet on
\ins{a} horizontal floor, and pushes \del{\eg} its hand on a wall in front of
it. The positive cone $\calC^+$ illustrates how the wall contact expands the
conventional support area between the two feet, while the negative cone
$\calC^-$ reveals that a second support area appears behind the two floor
contacts. This new area results from the ability for the robot to generate
downward forces at the wall contact.

Finally, let us determine in which cases a polygonal ZMP support area
can be found, which boils down to finding a suitable vector
$\bfn$.

\begin{proposition}
    \label{prop:finding-nemo}
    Either contact forces can generate any arbitrary resultant force or one can
    find a plane normal $\bfn$ such that the support area is polygonal.
\end{proposition}

\begin{IEEEproof}
Let $\calC_f$ denote the cone positively spanned by the
$\bff_i$'s. Then, the condition $\forall i, (\bfn \cdot \bff_i) > 0$
is equivalent to $\bfn\in\calC_f^*$, where $\calC_f^*$ is the
\emph{dual cone} of $\calC_f$ defined by
\begin{equation}
    \calC_f^* = \{\bfy :\ \forall \bfx\in\calC_f,\ \bfy\cdot\bfx \geq 0 \}.
\end{equation}
The set of solutions $\calC_f^*$ can be computed from $\calC_f$. In particular,
$\calC_f^*=\{\textbf{0}\}$ if and only if $\calC_f=\mathbb{R}^3$, \ie the
$\bff_i$'s positively span the whole space.
\end{IEEEproof}

Figure~\ref{fig:full-case} provides an example where contacts are in force
closure. As they can generate arbitrary contact wrenches, the support area
spans the whole virtual plane. \ins{Note that being able to generate arbitrary
forces is a} weaker\ins{ condition} than \emph{force closure}, which usually
assumes that contact forces can generate arbitrary resultant forces \emph{and
\del{torques}\ins{momenta}}.

\ins{Excluding such configurations, from Proposition~\ref{prop:finding-nemo}
one can choose $\bfn$ such that $\bfn \cdot \bff > 0$ for all resultant forces
$\bff$ generated by valid contacts. This is desirable insofar as it eliminates
potential singularities $\bfn \cdot \bff = 0$ where the ZMP would be
undefined~\eqref{eq:zmp}. In the classical horizontal-floor setting, taking
$\bfn$ as the floor normal is an example of such a solution.}

\del{Although historically the ZMP has been mostly used in locomotion, the full
support area constructed in this section is defined for contact wrenches in
general. As such, it is also applicable in related fields dealing with
\emph{mobility} under frictional constraints, such as grasping or workpiece
fixturing.}

\moved{As a mathematical object, support areas are 2D (non-linear) projections of the
6D contact wrench cone corresponding to the two components $\bfn \times
\bftau_O$ of the resultant moment. From there, one could question the
generality of this construction:~is it the ``best'' we can do? Could there
exist a 3D projection that would account for all three components of the
resultant moment? For the interested reader, we provide some elements of answer
to these questions in Appendix~\ref{app:better}.}

\begin{figure}
    \centering
    \includegraphics[height=5cm]{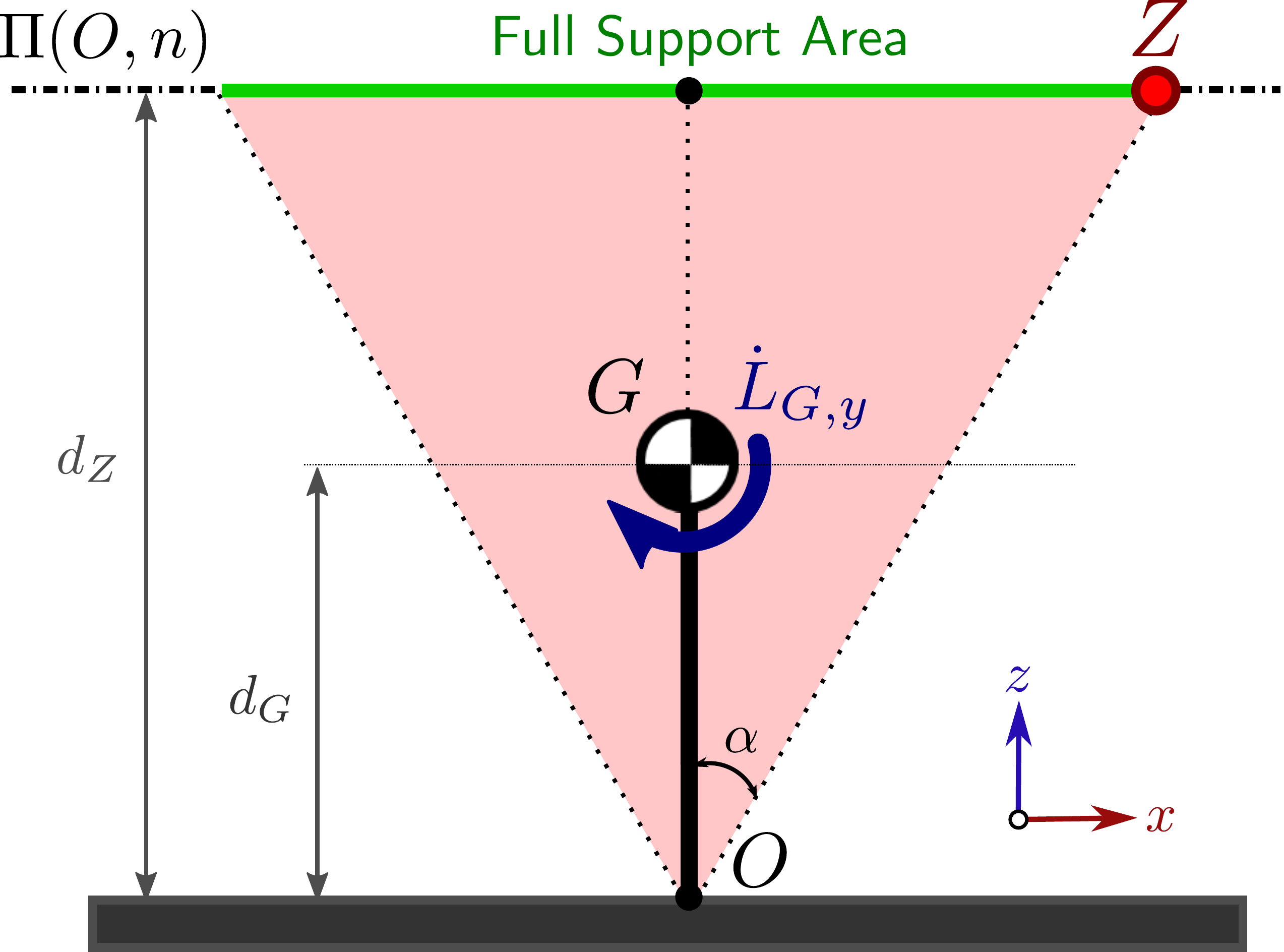}
    \caption{
        Illustration of the full support area (green) for a single point
        contact. Having the ZMP $Z$ (red) at the boundary of the support area
        generates an angular momentum $\dot{L}_{G,y}$ (blue) around the
        center-of-mass $G$ which increases with both the altitude $d_Z$ of the
        virtual plane and the proximity of $Z$ to the boundary of the support
        area.
    }
    \label{fig:farzmp}
\end{figure}

\subsection{Relation with the \ins{whole-body momentum}}

The full support area describes the ZMPs sustainable by valid contact forces.
As such, it does not take into account kinematic and dynamic limitations of the
robot itself, in particular the limited \del{range of wrenches }\ins{changes in
whole-body momentum }$(\dot{\bfP}, \LGd)$ that the robot can generate around
its COM by moving its limbs. Consider the simple example depicted in
Figure~\ref{fig:farzmp}, with a single point contact. According to
Proposition~\ref{prop:inter-poly}, the full support area is given by the
intersection of the friction cone with the virtual plane $\Pi(O, n)$. To
generate a ZMP $Z$, the robot needs to change its angular momentum around the
COM by
\begin{eqnarray}
    \dot{L}_{G,y} & = & - x_Z f_z + (d_Z - d_G) f_x
\end{eqnarray}
Thus, ZMPs close to the boundary of the full support area ($|x_Z| \to d_Z \sin
\alpha$) tend to generate higher angular momenta, a phenomenon which is
amplified by the altitude $d_Z$ of the virtual plane. In practice, legged
robots cannot generate a\del{n} arbitrary \del{dynamic wrenches around the COM}
\ins{angular momenta, as} torque limits and their bounded range of motion yield
additional constraints, which in turn restrict the ZMPs that the robot can
generate. One way to take these limitations into account is to put additional
constraints on the \ins{whole-body momentum}, as we will now see with the
linear pendulum mode.

\section{Pendular ZMP support area}
\label{sec:pendulum}

From a control point of view, the strong point of the ZMP lies in its direct
relationship with the acceleration of the COM. \ins{This can be seen with a
suitable rewriting of the Newton-Euler equations of the system:}

\begin{proposition}
    The COM acceleration, ZMP position and angular momentum are bound by the
    equation:
    \begin{equation}
        \label{eq:comdd}
        \pdd_G \ = \ \bfg 
        \ + \ \frac{\bfn \cdot (\pdd_G - \bfg)}{d_G - d_Z}\,\overrightarrow{ZG} \ + \ \frac{\bfn \times \LGd}{m (d_G - d_Z)},
    \end{equation}
    where $\bfg$ is the gravity vector. 
\end{proposition}

\begin{IEEEproof}
Expanding $\bftau_O = \bfp_G \times \bff + \LGd$ in
Equation~\eqref{eq:zmp-dZ},
\begin{align}
    (\bfn \cdot \bff) \bfp_Z &\,=\,\bfn \times (\bfp_G \times \bff) + \bfn
    \times \LGd + (\bfn \cdot \bfp_Z) \bff \\
    (\bfn \cdot \bff) \overrightarrow{GZ} &\,=\,(\bfn \cdot
    \overrightarrow{GZ}) \bff + \bfn \times \LGd.
\end{align}
\del{By definition of the gravito-inertial wrench, which is opposite to the
    contact wrench}\ins{Using Equations~\eqref{def-gi} and~\eqref{dyneq}},
    $\bff = m (\pdd_G - \bfg)$, so that:
\begin{equation}
    m (\bfn \cdot (\pdd_G - \bfg)) \overrightarrow{GZ} \ = \ m (\bfn \cdot
    \overrightarrow{GZ}) (\pdd_G - \bfg) + \bfn \times \LGd. \label{efgi}
\end{equation}
Equation~\eqref{eq:comdd} is a rearrangement of this last formula.
\end{IEEEproof}

\subsection{Linear pendulum mode}

The Linear Pendulum Mode (LPM) is a particular mode of
Equation~\eqref{eq:comdd}, obtained by regulating the COM altitude and \ins{the
}angular momentum to constant values:

\begin{definition}[Linear Pendulum Mode (LPM)]
    Provided with a normal vector $\bfn$, the linear-pendulum assumptions
    are:\footnote{
        Strictly speaking, our definition includes both $\bfn \times \LGd =
        \bm0$ and $\bfn \cdot \LGd = 0$, while only the former is necessary to
        realize the linear pendulum mode.
    }
    \begin{eqnarray}
        \label{lp-cons-1} \bfn \cdot \pdd_G & = & 0 \\
        \label{lp-cons-2} \LGd & = & \bm{0}.
    \end{eqnarray}
\end{definition}

When $d_Z < d_G$, the COM is \emph{above} the virtual plane with respect to the
plane normal. In such situations, Equation~\eqref{eq:comdd} yields the
well-known linear inverted pendulum mode (LIPM)~\cite{kajita2001iros,
sugihara2002icra}:

\begin{proposition}[Linear Inverted Pendulum Mode (LIPM)]
    Assume that $h \defeq d_G - d_Z > 0$, and define
    $\omega_{\mathrm{LIP}} =
    \sqrt{g / h}$. The COM acceleration in the \ins{LPM} is then related to the ZMP by:
    \begin{equation}
        \label{eq:comdd-lip}
        \pdd_G \ = \ \bfg - \omega_{\mathrm{LIP}}^2 \,\overrightarrow{GZ}
    \end{equation}
\end{proposition}

The LIPM was \del{taken for granted}\ins{a necessity} in previous works as the
ZMP was assumed to lie on the floor below the COM. But now that the virtual
plane coordinate $d_Z$ can be chosen freely, we may also consider the case
where $d_Z > d_G$, \ie taking the \del{COM \emph{below} the virtual
plane}\ins{virtual plane above the COM}. In this case,
Equation~\eqref{eq:comdd} yields a linear non-inverted pendulum\del{ mode}:

\begin{proposition}[Linear Non-inverted Pendulum Mode (LNPM)]
    Assume that $h' \defeq d_Z - d_G > 0$, and define
    $\omega_{\ins{\mathrm{LNP}}} =
    \sqrt{g / h'}$. The COM acceleration \ins{in the LPM} is then related to the ZMP
    by:
    \begin{equation}
        \label{eq:comdd-lp}
        \pdd_G \ = \ \bfg + \omega_{\ins{\mathrm{LNP}}}^2 \,\overrightarrow{GZ}
    \end{equation}
\end{proposition}

\del{The key difference between the two modes lies in the attractivity of the
ZMP.}\ins{The behavior of the ZMP with respect to COM acceleration differs
between the two modes.}
In the LIPM, the ZMP is a repulsor of the COM: when \ins{it} is maintained at a
fixed position, \ins{the COM} is ``pushed away'' from it and diverges to
infinity. \ins{When selecting a virtual plane below the center of mass $(d_Z <
d_G)$, ZMP trajectories will therefore follow the COM when it accelerates, and
precede it when it decelerates.}

In the \ins{LNPM}, the ZMP is a marginal attractor of the COM: when it is
maintained at a fixed position, the COM will ``orbit'' around it (yet without
asymptotic convergence, as there is no damping term in
Equation~\eqref{eq:comdd-lp}). \ins{In a virtual plane above the center of
mass ($d_Z > d_G$), ZMP trajectories will therefore precede the COM when it
accelerates, and follow it otherwise.}

\del{In this sense, the reactivity of a ZMP controller is less critical in the
LP mode:~the reference ZMP can stay fixed (with external perturbations
resulting in COM oscillations), whereas it needs to be constantly moving in the
LIP mode (where uncorrected perturbations result in COM divergence).}

\ins{These two complementary behaviors reflect the fact that, in the LPM, the
resultant contact force $\bff$ always points from the COM to the ZMP.} In what
follows, we \del{will}\ins{choose to} take the virtual plane above the center
of mass \del{($d_Z > d_G$) }and use the LNPM.

\subsection{Shrinking of the support area}

Previous methods~\cite{kagami2002ar, sugihara2002icra, kajita2003icra,
harada2006tro, shibuya2006iecon, sato2011tie} applied the LIPM using the convex
hull of contact points (CHCP) as ZMP support area. By
Proposition~\ref{prop:inter-poly}, \ins{when all contacts are coplanar} the
CHCP is the full support area. However, it turns out that the
constraints~\eqref{lp-cons-1}-\eqref{lp-cons-2} of the linear-pendulum mode
shrink the ZMP support area.

\begin{figure}
    \centering
    \includegraphics[trim={1mm 0 0 2mm},clip,width=0.98 \columnwidth]{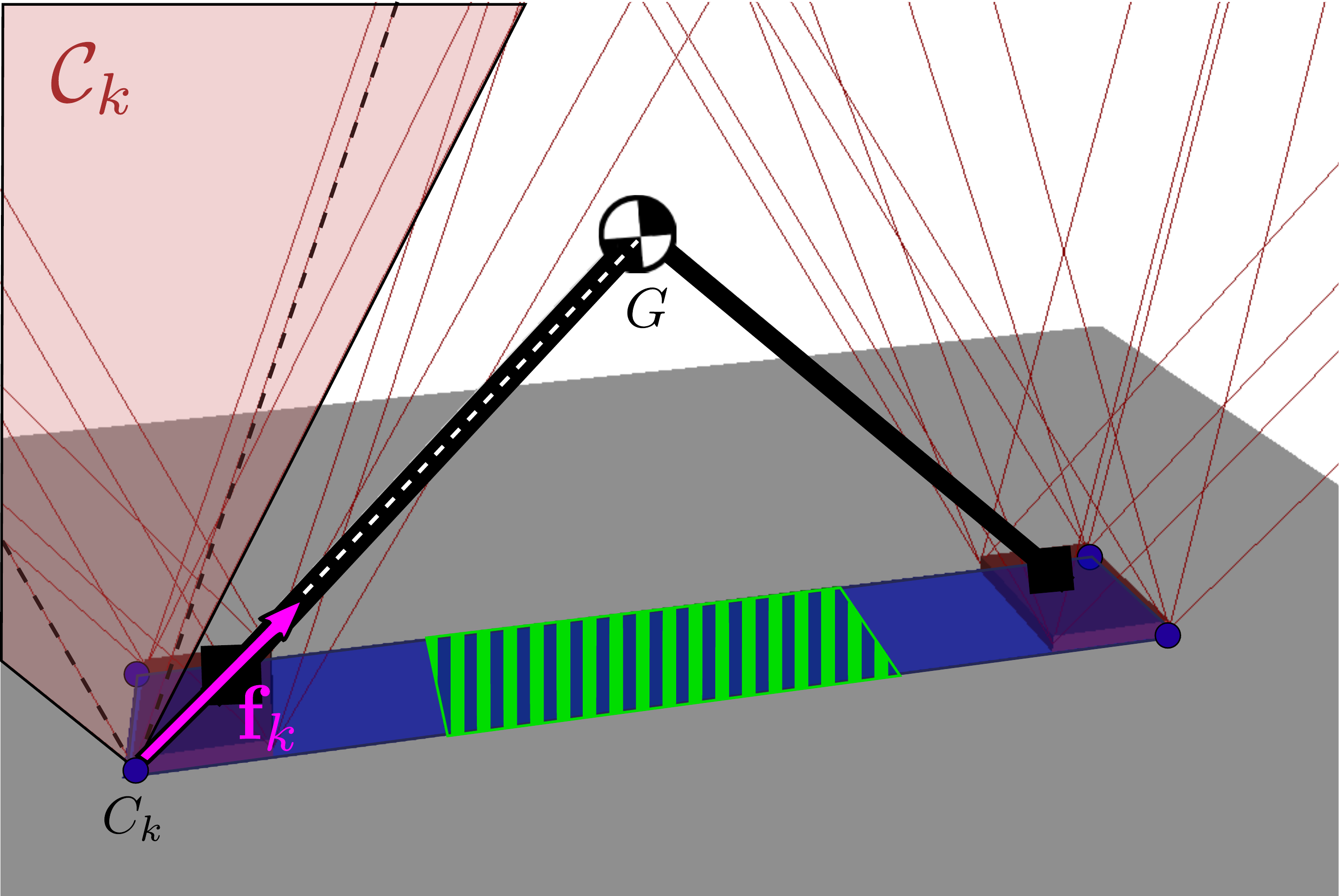}
    \caption{
        Situation where the pendular support area (green stripes), \ie the ZMP
        support area for the Linear Pendulum Mode, is smaller than the convex
        hull of ground contact points (blue polygon). The robot has its legs
        stretched, with its two feet one meter apart on a horizontal floor (in
        gray). Its COM is 50 cm above ground, and the friction at contact is
        $\mu=0.5$. The ZMP cannot be located at the corner $C_k$ of the convex
        hull, as it would require a resultant contact force $\bff_k$ (in
        magenta) lying outside of the friction cone $\calC_k$ (in red). The
        pendular support area in this Figure has been computed using the
        algorithm from Section~\ref{sec:algo-cdd}.
    }
    \label{fig:chcp}
\end{figure}

This can be seen as follows. Suppose that the ZMP is located at a vertex $C_k$
of a given contact polygon. The resultant contact force $\bff$ must then be
realized as one contact force $\bff_k$ applied at $C_k$, while all other
contact forces $\bff_j = \bm0$ for $j \neq k$. (From Equation~\eqref{eq:zcomb},
$\bfp_Z = \bfp_{Z_k} = \bfp_{C_k}$ implies that only the $\lambda_i$'s
corresponding to $C_k$ can be strictly positive.) However, in the LPM the
contact force $\bff = \pdd_G - \bfg$ is also directed from $Z=C_k$ to $G$.
Hence, the ZMP support area cannot be the CHCP as soon as the COM lies outside
of the friction cone of at least one ground contact point. An example of such
situation is when a biped robot stretch\ins{es} its legs, as depicted in
Figure~\ref{fig:chcp}.

Methods that take the Convex Hull of Contact Points (CHCP) as ZMP support areas
rely by construction on the assumption of ``infinite friction'' to rule out
such cases, at the cost of being problematic for locomotion on low-friction
floors~(apart from sliding, foot yaw rotations due to insufficient friction
have also been observed and studied~\cite{cisneros2014icma, caron2015icra}).
\del{To the best of our knowledge,} This \ins{assumption explains why the
}shrinking of the ZMP support area in the LPM was \del{previously
unnoticed}\ins{not highlighted in previous works}.

We call \emph{pendular support area} the shrunk support area that takes into
consideration the Equations~\eqref{lp-cons-1}-\eqref{lp-cons-2} of the LPM.

\subsection{Computation of the \ins{pendular} support area}
\label{sec:algo-cdd}

In this section, we assume that $\bfn = \bfe_Z$, the upward vertical (opposite
to gravity) unit vector. We now propose an algorithm to compute the pendular
support area based on the double-description method~\cite{fukuda1996double}. It
turns out that computations of this set are very similar to that of the COM
static-\del{stability}\ins{equilibrium} polygon. We explain both calculations here.

\subsubsection{Static-equilibrium polygon} In static equilibrium, the equations
on the contact wrench are:
\begin{eqnarray}
    \bff & = & -m \bfg \\
    \label{eq:ss-2} \bfn \times \bftau_O & = & -\bfn \times (\bfp_G \times m\bfg) \\
    \bfn \cdot \bftau_O & = & 0
\end{eqnarray}
By expanding the triple product in Equation~\eqref{eq:ss-2}, we can rewrite
them equivalently as:
\begin{equation}
    \label{ss-poly-1}
    \left[
        \begin{array}{cc}
            \bfE_3 & \bm{0}_{3 \times 3} \\
            \bm{0}_{1 \times 3} & \bfn^\top
        \end{array}
    \right] 
    \left[
        \begin{array}{c}
            \bff \\
            \bftau_O
        \end{array}
    \right] 
    \ = \ 
    \left[
        \begin{array}{c}
            -m \bfg \\
            0
        \end{array}
    \right]
\end{equation}
\begin{equation}
    \label{ss-poly-2}
    \bfp_G \ = \ (\bfn / m g) \times \bftau_O + z_G \bfn
\end{equation}
In concise form, these two equations can be written:
\begin{eqnarray}
    \bfA \bfw_O & = & \bfb \\
    \bfp_G & = & \bfC \bfw_O + \bfd
\end{eqnarray}

Consider the stacked vector of contact forces $\bff_{all} = [ \bff_1^\top \cdots
\bff_n^\top ]^\top$. Linearized friction cones are given by linear inequalities
$\bfF_i \bff_i \leq \bm{0}$. For instance, four-sided friction pyramids can
be formulated as
\begin{equation}
    \bfF_i \ = \ \left[
        \begin{array}{ccc}
            -1 & 0 & - \mu_i \\
            +1 & 0 & - \mu_i \\
            0 & -1 & - \mu_i \\
            0 & +1 & - \mu_i
        \end{array}
    \right] \bfR_i^\top.
\end{equation}
Combining all $\bfF_i$'s in a block diagonal matrix $\bfF$ yields an inequality
$\bfF \bff_{all} \leq \bm{0}$. Meanwhile, Equation~\eqref{def-c} provide\ins{s} a
linear mapping $\bfw_O = \bfG_O \bff_{all}$ from contact forces to the contact
wrench, where $\bfG_O$ is the so-called \emph{grasp matrix}. \del{Summing
up}\ins{Then}, the
set of \ins{valid} contact forces in static\del{-stability}\ins{ equilibrium} is
given by:
\begin{eqnarray}
    \bfF \bff_{all} & \leq & \bm{0} \\
    \bfA \bfG_O \bff_{all} & = & \bfb
\end{eqnarray}
This expression of a polytope by linear inequalities is known as the
\emph{half-space} representation. By the Minkowski-Weyl
theorem~\cite{fukuda1996double}, all polytopes can be equivalently written in
terms of linear inequalities (half-space representation) or in terms of
vertices and rays (span representation). The double description
method~\cite{fukuda1996double} provides a ``black-box'' algorithm to convert
between one representation and the other. Using this tool, one can compute the
span representation of this set as:
\begin{equation}
    \bff_{all} = \sum_i \alpha_i \bfv_i
\end{equation}
where $\alpha_i > 0, \sum_i \alpha_i=1$, and the $\bfv_i$'s are computed
vertices. The span representation of the stability polygon is finally given by
\begin{equation}
    \bfp_G = \sum_i \alpha_i (\bfC \bfG_O \bfv_i) + \bfd.
\end{equation}
In the case of static stability, the solution is always a polygon, so there is
no need to consider rays in this span representation. Note that this method for
computing the COM static-\del{stability}\ins{equilibrium} polygon by
double-description is different from the recursive polytope projection
algorithm~\cite{bretl2008tro}.

\subsubsection{Pendular support area} the
Equations~\eqref{lp-cons-1}-\eqref{lp-cons-2} of the LPM are written, in
terms of the contact wrench:
\begin{eqnarray}
    \label{lp-wrench1} \bfn \cdot \bff & = & mg \\
    \label{lp-wrench2} \bfn \cdot \bftau_G & = & 0 \\
    \label{lp-wrench3} \bfn \times \bftau_G & = & \bm{0}
\end{eqnarray}
where $g \approx 9.81$ m.s$^{-2}$ is the gravity constant. Taking the resultant
moment $\bftau_O$ rather than $\bftau_G$, one can rewrite the equations above
as:
\begin{equation}
    \label{eq:zmp-sa-1}
    \frac{1}{mg}
    \left[ 
        \begin{array}{cc} 
            z_G \bfE_3 & [\bfn \times] \\ 
            -(\bfn \times \bfp_G)^\top & \bfn^\top
        \end{array}
    \right] 
    \left[
        \begin{array}{c}
            \bff \\
            \bftau_O
        \end{array}
    \right] 
    \ = \ 
    \left[
        \begin{array}{c}
            \bfp_G \\
            0 \\
        \end{array}
    \right]
\end{equation}
\begin{equation}
    \label{eq:zmp-sa-2}
    \bfp_Z \ = \ \frac{z_Z - z_G}{mg} \bff + \bfp_G,
\end{equation}
Equation~\eqref{eq:zmp-sa-1} results from
\eqref{lp-wrench1}-\eqref{lp-wrench2}, while Equation~\eqref{eq:zmp-sa-2} is a
reformulation of \eqref{lp-wrench3}.

Contrary to the previous setting, where the COM position $\bfp_G$ resulted from
the contact wrench in static equilibrium, we now assume that $\bfp_G$ is known
(\eg measured from the instantaneous robot state). In concise form, the two
equations above can be written:
\begin{eqnarray}
    \bfA' \bfw_O & = & \bfb' \\
    \bfp_Z & = & \bfC' \bfw_O + \bfd'
\end{eqnarray}
where the matrices $\bfA', \bfC'$ and vectors $\bfb', \bfd'$ now depend on
$\bfp_G$. From there, the computations are the same as for static stability:
the set of \ins{valid} contact forces $\bff_{all}$ is given by
\begin{eqnarray}
    \bfF \bff_{all} & \leq & \bm{0} \\
    \bfA' \bfG_O \bff_{all} & = & \bfb'
\end{eqnarray}
Using the double description method, one can compute the span representation
of this set as 
\begin{equation}
    \bff_{all} = \sum_i \alpha_i \bfv_i + \sum_j \lambda_j \bfr_j
\end{equation}
where $\alpha_i > 0, \sum_i \alpha_i=1$, $\lambda_j > 0$ and the $\bfv_i$'s
(rep. $\bfr_j$'s) are the computed vertices (resp. rays). The span
representation of the pendular support area is finally given by
\begin{equation}
    \bfp_Z = \sum_i \alpha_i (\bfC' \bfG_O \bfv_i) + \sum_j \lambda_j (\bfC' \bfG_O  \bfr_j) + \bfd'.
\end{equation}

Similarly to the full support area, the pendular area can be either conical or
polygonal. Compared with the full support area or static\del{
    stability}\ins{-equilibrium} polygon, it depends not only on the set of
contacts $\{ \bfp_{C_i} \}$, but also on the instantaneous position $\bfp_G$ of
the COM. Yet, where the full support area yields only a necessary condition for
contact stability (it is a 2D projection of the 6D contact wrench, the
remaining four components being unconstrained), the pendular support area gives
a condition both necessary \emph{and sufficient} under the LPM (the four other
components of the contact wrench being determined by equality constraints).
        
\section{Experiments}
\label{sec:experiments}

\subsection{Trajectory generation}
\label{sec:trajgen}

We now design a trajectory generator for ZMP-COM trajectories based on the
model-preview control formalism introduced in previous
works~\cite{kajita2003icra, audren2014iros}. We use the ZMP as a command and
COM as the output variable. First, we \del{interpolate a}\ins{define the
support of the} ZMP trajectory $\bfp_Z(\gamma(t))$ as a line segment:
\begin{equation}
    \bfp_Z(\gamma(t)) \ = \ \gamma(t) \bfp_1 + (1 - \gamma(t)) \bfp_0,
\end{equation}
where $\bfp_0$ (resp. $\bfp_1$) denotes the initial (resp. final) ZMP position.
\ins{Assuming that its initial velocity is zero, t}he COM \del{then }follows a
parallel line segment $\bfp_G(\eta(t))$:
\begin{equation}
    \bfp_G(\eta(t)) \ = \ \eta(t) \bfp_1 + (1 - \eta(t)) \bfp_0,
\end{equation}
where $\ddot{\eta}(t) = \omega^2 (\gamma(t) - \eta(t))$.
\ins{Next, d}efine the state of the control problem by $\bfx(t)~=~[ \eta(t) \ \dot{\eta}(t)
]^\top$ and its command by the linear position $\gamma(t)$ of the ZMP.
Discretizing the time interval into $K$ steps of duration $\delta t$, the
system's \del{linear }dynamics become
\begin{eqnarray}
    \label{lindyn}
    \bfx_{k+1} & = &
    \left[
        \begin{array}{cc}
            \cos(\omega \delta t) & \frac{1}{\omega} \sin(\omega \delta t) \\
            - \omega \sin(\omega \delta t) & \cos(\omega \delta t)
        \end{array}
    \right]
    \bfx_k
    \\ & + &
    \nonumber
    \left[
        \begin{array}{c}
            1 - \cos(\omega \delta t) \\
            \omega \sin(\omega \delta t)
        \end{array}
    \right]
    \gamma_k
\end{eqnarray}
Let $\bfX = [\bfx_0^\top \cdots \bfx_K^\top]^\top$ and $\bfgamma = [\gamma_0 \cdots
\gamma_{K-1}]^\top$. Applying \eqref{lindyn} repeatedly, we build the matrices
$\bfPhi$ and $\bfPsi$ such that $\bfX = \bfPhi \bfx_0 + \bfPsi \bfgamma$. We assume
that the system starts with zero COM velocity, so that $\bfx_0 = \bm{0}$ and
$\bfX = \bfPsi \bfgamma$.

\begin{figure}[t]
    \centering
    \includegraphics[height=4.9cm]{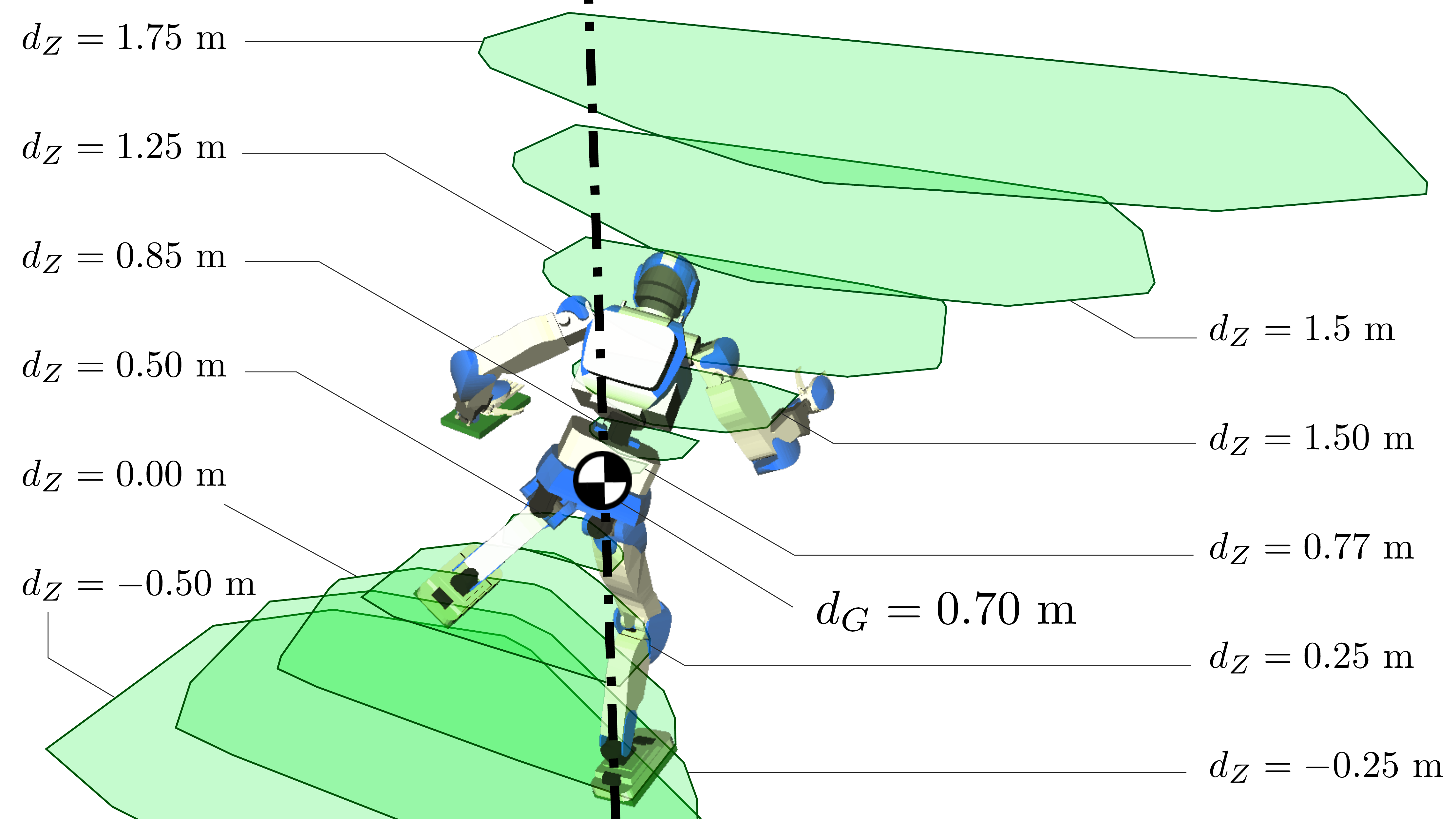}
    \caption{
        Variations of the shape of the pendular support area with the
        coordinate $d_Z$ of the virtual plane. In this configuration, the
        humanoid is making two tilted foot contacts and a left hand contact.
        (There is no right hand contact.) The axis directed by the plane normal
        $\bfn$ and going through the COM is depicted by a dashed line. In the
        proposed method, the plane altitude can be chosen freely as long as
        $d_Z \neq d_G$, however the support area becomes small when $d_Z$ and
        $d_G$ are close.
    }
    \label{fig:pendslice}
\end{figure}

We formulate the trajectory generation problem as a Quadratic Program (QP) as
follows:
\begin{align*}
    \textbf{Objective:} \quad & \min \ w_1 c_1(\bfgamma) + w_2 c_2(\bfgamma) \\
    \textbf{Constraints:} \quad & \bm{0} \leq \bfgamma \leq \bm{1} \\
                                & \bfx_K = \bfPsi_{\mathrm{last}} \bfgamma = [\:1\ 0\:]^\top \\
                                & \bfgamma_{K-1} = 1
\end{align*}

The objective is the weighted sum of two terms:
\begin{eqnarray}
    c_1(\bfgamma) & = & \textstyle \frac1K \sum_k (\eta_k - \gamma_k)^2 \\
    c_2(\bfgamma) & = & \textstyle \sum_k (\gamma_k - \gamma_{k-1})^2
\end{eqnarray}
The first one minimizes COM accelerations while the second regularizes the ZMP
trajectory. The weights of the cost function are set to $w_1 = 1$ and $w_2 =
100$. Meanwhile, the constraints ensure respectively that:
\begin{itemize}
    \item the ZMP belongs to the line segment $(\gamma(t) \in [0, 1])$
    \item the COM ends at the destination point $(\eta_K = 1)$ with zero
        velocity $(\dot{\eta}_K = 0)$,
    \item the ZMP also ends at the destination point $(\gamma_{K-1} = 1)$.
\end{itemize}
\ins{
    The solution to this QP provides a ZMP trajectory $p_Z(t)$, which is then
    integrated into a COM trajectory $p_G(t)$ by applying~\eqref{eq:comdd-lp}.
    We also added a damping term at this stage to smooth out undesired
    COM oscillations. On a side note, }set aside the two regularization objectives, this
optimization problem falls under the framework of Time-Optimal Path
Parameterization (TOPP)~\cite{Pha14tro, PS15tmech}. One could then trade smoothness of COM
accelerations for Admissible Velocity Propagation (AVP), allowing for the
integration into a kinodynamic planner of COM trajectories~\cite{PhaX13rss}.

\begin{figure}[t]
    \centering
    \includegraphics[height=4.9cm]{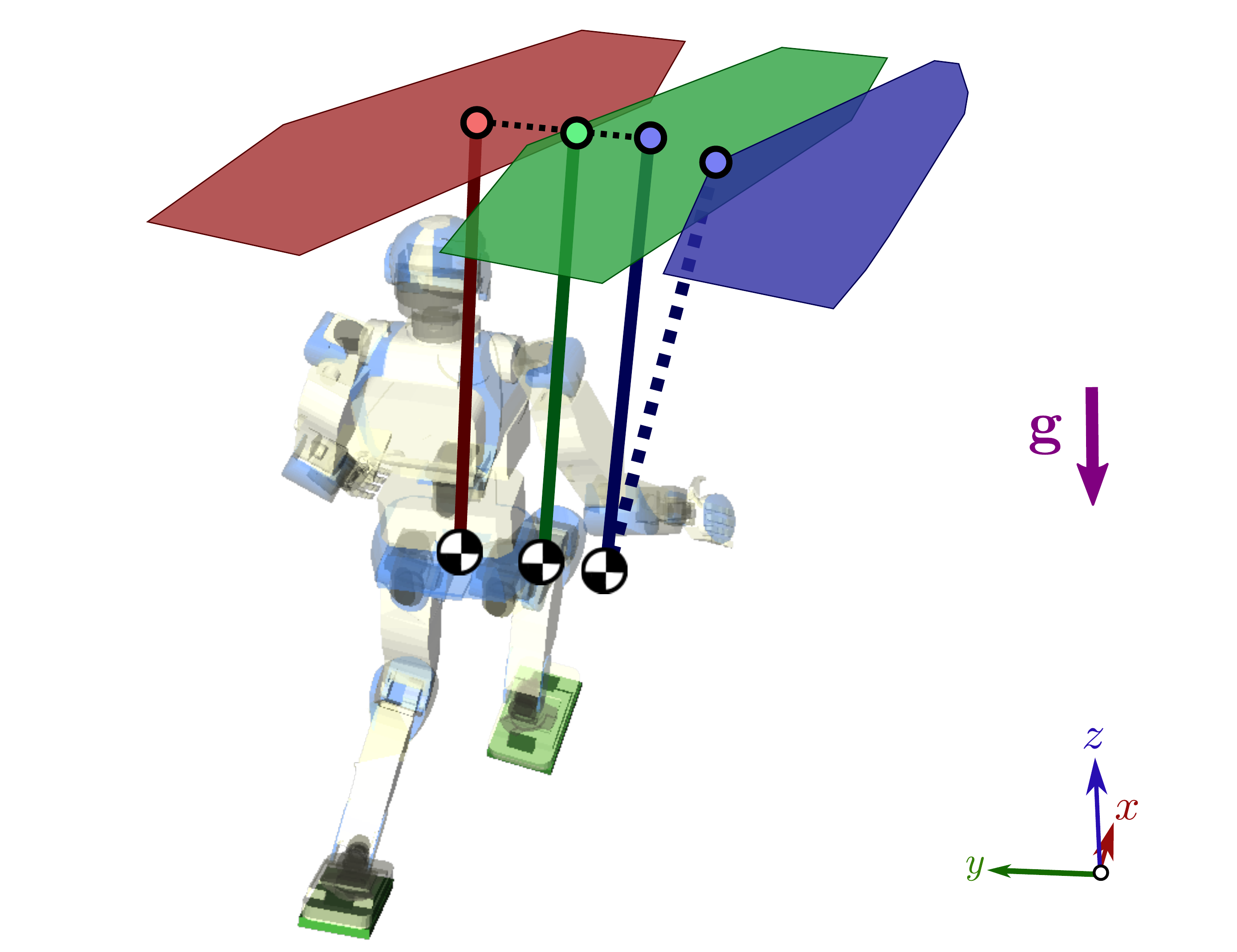}
    \caption{
        Three pendular support areas and their corresponding pendulums at rest
        configurations, \ie with the ZMP at the vertical of the COM.
        \del{Colors match areas with their respective pendulums.} The
        \del{transparent }HRP-4\ins{ posture} corresponds to the leftmost (red)
        configuration\del{. In this case, the COM lies between contacts and the
        ZMP }\ins{, where the vertical of the COM} is centered in the area. In
        the middle (green) case, the COM slid to the right while the area slid
        by a lesser amount, so that the vertical of the COM is now on its edge:
        the configuration is statically marginally stable. In the rightmost
        (blue) case, the pendulum slid further: the vertical of the COM is no
        longer in the support area, and the leftmost feasible ZMP (dashed line)
        will steer the COM further away to the right.
    }
    \label{fig:pendaway}
\end{figure}

With this method, contact stability is mostly enforced by checking
that the segment $[\bfp_0, \bfp_1]$ lies inside the pendular support area
computed for the extremities $\bfp_0$ and $\bfp_1$ of the segment. However, the
COM \del{will} move\ins{s} as the robot performs the motion, which affects both
the position and shape of the pendular support area. This phenomenon can have
two undesired outcomes:
\begin{itemize}
    \item the area becomes empty: we observed empirically that this would only
        happen when the COM is \del{very }far away (\eg more than one meter) from
        contacts, and was not a threat in practice.
    \item the area constraints the COM motion toward divergence:~past a certain
        COM-to-contacts distance, the support area slants away from contacts,
        in a way such that it contains no ZMP that could bring the pendulum
        back above contacts.
\end{itemize}
Figure~\ref{fig:pendaway} illustrates this phenomenon. In practice, ensuring
that the ZMP is well within the support area computed at the beginning $\bfp_0$
and end $\bfp_1$ of the segment was enough to rule out both undesirable
outcomes.

\subsection{Long stride while leaning on a wall}

\begin{figure}[t]
    \centering
    \includegraphics[width=0.98\columnwidth]{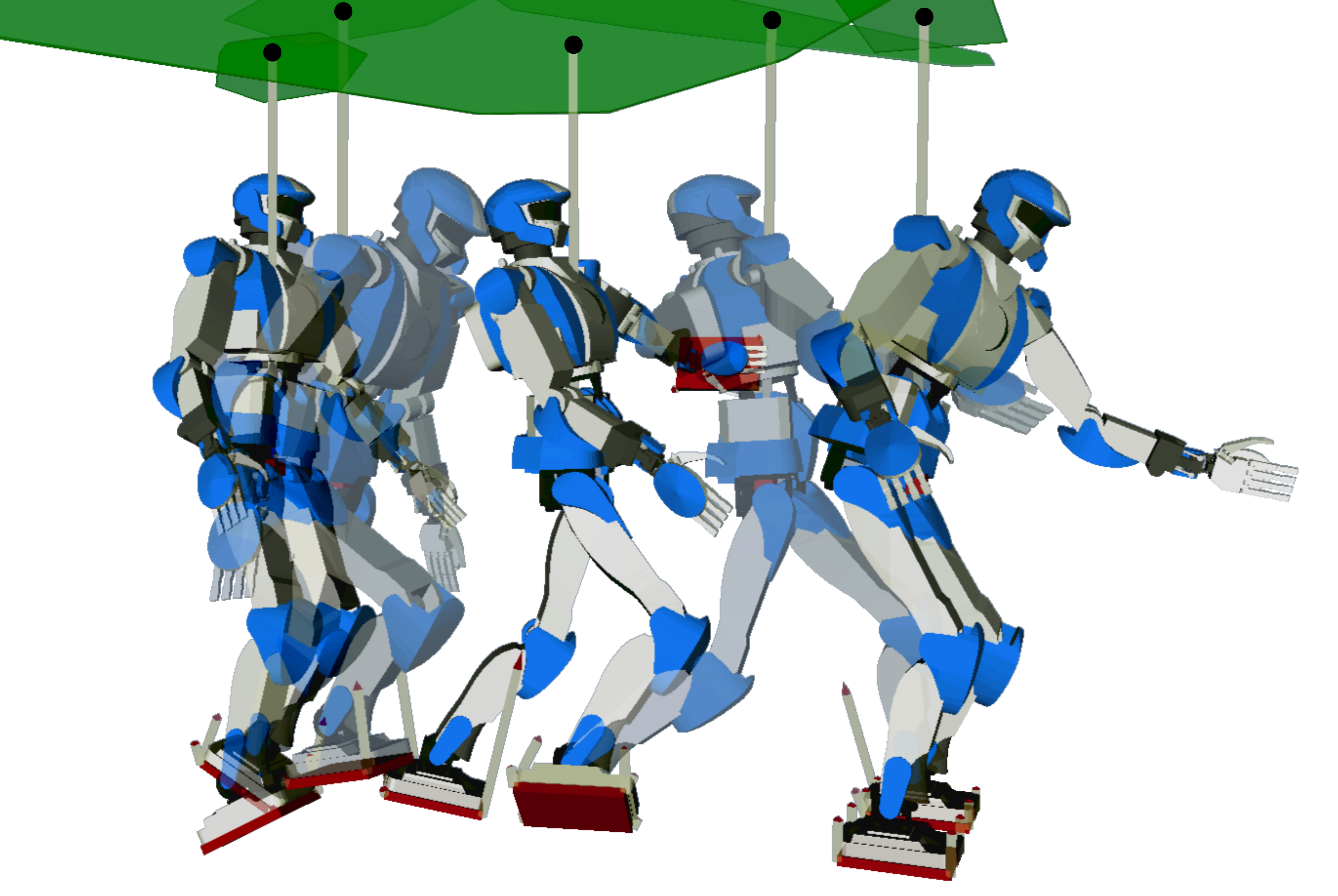}
    \caption{
        Snapshots of the motion generated \del{by LP control}\ins{in the Linear
        Pendulum Mode} with a ZMP above the robot's COM. \del{The scenario is
        designed so that} The robot has to put its left hand on a ``wall''
        contact (in the background) and its right foot on the opposite tilted
        platform in order to perform an ample swing of the left leg that is
        otherwise impossible. In these simulations, the virtual plane is taken
        one meter above the robot's COM. Green polygons in th\ins{is virtual}
        plane \del{above the
        robot's head are the respective}\ins{correspond to the} pendular
        support areas for each snapshot. \del{The (virtual) gray wires connect ZMPs
        to their respective COMs.}\ins{Vertical lines represent the (virtual)
        linear pendulum.} Contact forces (arrows pointing from contacts
        toward the robot) are computed at each time instant to cross-validate
        \del{our}\ins{the} contact-stability \del{criterion}\ins{of the motion}.
    }
    \label{fig:motion}
\end{figure}

We implemented the whole pipeline described so far to generate multi-contact
motions for a model of the HRP-4 humanoid robot. The scenario is depicted in
Figure~\ref{fig:motion}. The robot has to step on inclined platforms in order
to reach its goal configuration on the right. Because there is no platform for
its left foot in the middle of the course, the only way for it to complete the
task is to \del{use of}\ins{put its left hand on} the elevated ``wall''
platform \del{with its left hand }while \del{keeping}\ins{pushing with} its
right foot on the opposite tilted surface. Relying on these two simultaneous
contacts, the humanoid can perform a long stride\del{ with its left leg,} which
would have been impossible to achieve in single-support.

As input given to solve this scenario, we assume that a contact planner
provides a sequence of \emph{stances}, where a stance $\sigma_i (\{ \bfp_{C_i}
\}, \bfp_i)$ provides both a reference position of the COM $\bfp_i$ and a set
of contact points $\{ \bfp_{C_i} \}$. The first stage of our solution computes
stance-to-stance COM trajectories. To move from $\sigma_i$ to stance
$\sigma_{i+1}$, the controller considers the line segment $[\bfp_i,
\bfp_{i+1}]$. The trajectory generator is called if this segment is included in
the pendular support area $\calS(\bfp_i, d_Z)$ for the initial COM position.
Otherwise, $d_Z$ is increased until the segment is included in $\calS(\bfp_i,
d_Z)$. This condition is easy to fulfill in practice, as we observed that the
region $\calS(\bfp_G, d_Z)$ grows like the section by the plane $\Pi(d_Z,
\bfn)$ of a cone passing through $G$ (see Figure~\ref{fig:pendslice}). In the
motion depicted in Figure~\ref{fig:motion}, this process lead us to take the
virtual plane one meter above the center of mass of the humanoid (HRP-4 is
1.5-meter tall).
\del{Note that, }Although contact planning was done by hand in this experiment,
we \ins{also }used the pendular support area to select COM positions $\bfp_i$
from contact locations $\{ \bfp_{C_i} \}$, by enforcing that the ZMP at the
vertical of $\bfp_i$ lies well inside $\calS(\bfp_i, d_Z)$.\del{ As such, this
area can also be used to compute stances from sets of contacts.}

Once a reference COM trajectory $\bfp_G(t)$ has been computed, we generate
whole-body joint-angles by differential inverse kinematics (IK) under the
following constraints, by decreasing task weight:
\begin{enumerate}
    \item tracking of contacting end-effector poses, 
    \item tracking of COM trajectory, 
    \item tracking of free end-effector poses, 
    \item minimum variations in angular momentum
    \item preferred values for some joint-angles. 
\end{enumerate}
We \del{used}\ins{applied} our own IK solver for the task, which is
\del{published}\ins{released} in the \textit{pymanoid}
library.\footnote{\url{https://github.com/stephane-caron/pymanoid}} Similarly
to \cite{lee2010iros}, this solver relies on a single-layer QP problem\ins{,
using the above cost function and a set of inequality constraints
\begin{equation}
    \qd\ \leq\ \min(\qd_{\textrm{max}}, -K_s (\bfq - \bfq_{\textrm{max}}))
\end{equation}
to limit joint velocities}. Gains and weights used in the simulations are
reported in Table~\ref{table:ik}, while other simulation parameters are given
in Table~\ref{table:sims}. \ins{Implementation details can be checked in the
library and source code released with the paper\ins{ (see the
\textit{motion\_editor} distributed in~\cite{code})}.}

\begin{figure}[t]
    \centering
    \includegraphics[width=0.98\columnwidth]{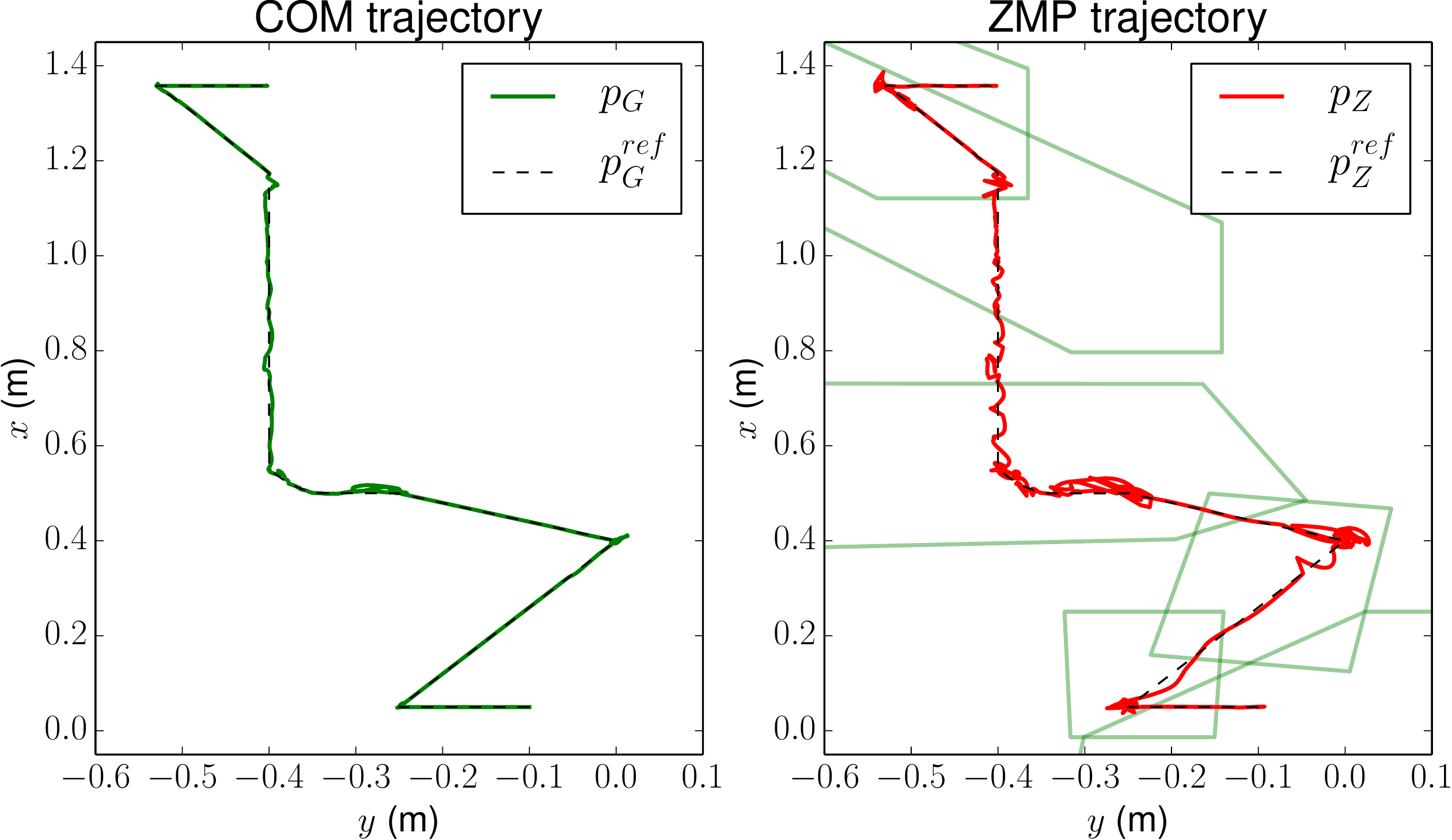}
    \caption{
        \ins{COM and ZMP trajectories (resp. in green and red) for the motion of
        Figure~\ref{fig:motion}. Reference trajectories are straight dashed
        lines between via points. Five pendular support areas corresponding to
        single- and double-support configurations are plotted for reference
        (green polygons). The accompanying video~\cite{code} shows the
        evolution of support areas throughout the motion.}
    }
    \label{fig:2dtraj}
\end{figure}

ZMP tracking is not an explicit task in the list above, and is realized as a
side effect of the COM tracking task. The latter comes second in the hierarchy
of the differential IK solver and is therefore not fulfilled perfectly, as
illustrated in Figure~\ref{fig:2dtraj}. The same holds for the task $\LGd = 0$,
which has an even lower rank in the hierarchy. Yet, ZMP deviations in the
generated trajectory are small enough and one can check that the ZMP always
stays well within the pendular support area (see the accompanying video).
\del{Furthermore, }\ins{W}e cross-validated the \del{dynamic}
\ins{contact-}stability of \del{our generated}\ins{the final} motion by
computing, at each time step, a set of valid contact forces $\bff_{all}$. Force
vectors are depicted in Figure~\ref{fig:motion} and displayed in the
accompanying video~\cite{code}.

\ins{\subsection{Computation times}}

\begin{table}[t]
    \caption{
        Computation times (in ms) for the full support area and
        CWC~\cite{caron2015rss}, as well as for the pendular support area
        using~\cite{fukuda1996double} (\textsc{cdd}) or~\cite{bretl2008tro}
        (B\&L). Bold indicates fastest in each category.
    }
    \label{table:times}
    \centering
    \begin{tabular}{|c||c|c|c|}
        \hline
        Stability Criterion & One contact & Two contacts & Three contacts$^6$ \\
        \hline
        \hline
        CWC (\textsc{cdd}) & $1.1 \pm 4.2$ & $4.5 \pm 1.8$ & $9.5 \pm 2.6$ \\
        \hline
        Full support area & $\bf 0.8 \pm 0.4$ & $\bf 2.6 \pm 2.1$ & $\bf 6.4 \pm 3.1$ \\
        \hline
        \hline
        Pendular area (\textsc{cdd}) & $\bf 6.4 \pm 3.3$ & $\bf 24.8 \pm 4.7$ & $385.3 \pm 251.3$ \\
        \hline
        Pendular area (B\&L) & $18.5 \pm 3.7$ & $33.9 \pm 12.2$ & $\bf 67.9 \pm 11.9$ \\
        \hline
    \end{tabular}
\end{table}

\ins{
    Table~\ref{table:times} reports average computation times for both kinds of
    support area during the execution of the motion from
    Figure~\ref{fig:motion}. Simulations were run on an average laptop computer
    with an Intel\textsc{(r)} Core\textsc{(tm)} i7-6500U CPU @ 2.50 Ghz.
    \ins{The source code for these simulations is located in the
    \textit{full\_support\_area} folder of~\cite{code}.}

    \ins{The number of samples for each average in Table~\ref{table:times} is
    around 1000 for one- and two-contact configurations, and around 100 for
    three-contact configurations.}
    The first two lines show that computing the full support area is faster
    using our geometric construction than by projecting the
    CWC~\cite{caron2015rss}, which is expected since the latter is a more
    complex 6D cone. These two criteria are always computed faster than a
    pendular support area, as the latter adds equality constraints to the
    former, while neither of them takes into account limits on the angular
    momentum. Also, recall that the full support area is only a necessary
    condition for contact stability.

    The first line of the second group corresponds to the algorithm described
    in Section~\ref{sec:algo-cdd}. We also implemented a variant of the
    projection algorithm from~\cite{bretl2008tro} to compute the pendular
    support area.\footnote{ See the \textit{contact\_stability} ROS package
    in~\cite{code} for details.} In practice, this variant is one to three
    times slower than \textsc{cdd}, except for triple contacts where we
    observed a significant slowdown of \textsc{cdd}, along with some freezes.
    This point is implementation-specific rather than a
    limitation of the underlying double-description algorithm.\footnote{
        Averages for ``Pendular area (\textsc{cdd})'' in triple-contact are
        reported for samples where \textsc{cdd} did not freeze, amounting for
        around $20\%$ of all samples.} 
}

\section{Conclusion and future work}
\label{sec:conclusion}

In this paper, we have generalized the notion of ZMP \emph{support areas} to
take into account frictional constraints and multiple non-coplanar contacts.
First, we derived the geometric construction of the \emph{full support area},
which contains all ZMPs generated by valid contact forces. Then, we moved to
the control of humanoid robots \del{with}\ins{in} the Linear Pendulum Mode\del{
    (LPM)}. We noticed that the constraints stemming from this mode shrink the
support area (even on horizontal floors\del{, although it was previously
unnoticed}), and proposed an algorithm to compute the new area, which we called
\emph{pendular support area}. Armed with these new conceptual tools, we
designed a whole-body controller for locomotion across arbitrary multi-contact
stances, which we demonstrated in simulation with a model of the HRP-4 humanoid
robot.

Support areas are a general property of contact wrenches\del{ with no
particular ties to the problem of locomotion. As a result, they also have
potential applications in related fields dealing with }\ins{ and can be applied
in all fields related to }\emph{mobility}, \del{such as}\ins{including
locomotion,} grasping or workpiece fixturing. As a mathematical object,
\del{support areas}\ins{they} are 2D \del{(}non-linear\del{)} projections of
the 6D contact wrench cone\del{corresponding to the two components $\bfn \times
\bftau_O$ of the resultant moment}. From there, one \del{could question the
generality of this construction}\ins{may ask}:~is it the
\del{``best''}\ins{most general} we can do\del{? Could there exist }\ins{, or can }a
3D projection\del{ that would}\ins{ take into} account \del{for all
three}\ins{more} components of the resultant moment? For the interested reader,
we provide some elements of answer \del{to these questions }in
Appendix~\ref{app:better}\ins{, although the question itself remains open}.

\ins{In practice, w}e believe that the ability to take ZMPs in virtual support
areas paves the way for multiple future developments. \ins{For contact
planning,} we noticed how individual \del{contact}\ins{support} polygons can be
used \del{by a contact planner }to measure the ``increase of stability'' of a
prospective contact. 
\ins{
    For whole-body control, our approach can be followed in the more general
    framework of \emph{a-priori} dimensionality reduction of control problems.
    At one end of the spectrum, the full support area corresponds to a
    controller with state variables for all six components of the whole-body
    momentum, such as \cite{nagasaka2012rs, audren2014iros}. The task of such a
    controller is harder, but the area depends only on contact locations. At
    the other end of the spectrum, the pendular support area corresponds to
    fixing four momentum components, thus reducing control to a two-dimensional
    problem. The task of the controller is then simpler, at the cost that the
    area depends on the position of the COM. Between these two ends of the
    spectrum lies a hierarchy of intermediate three-, four- or five-dimensional
    problems that can be computed, in a similar fashion, by a-priori reduction
    of equality constraints. Exploring these possibilities is the focus of our
    current research.
}

\del{Finally, from a neurophysiological perspective, the control of the COM
\emph{from above} proposed in this paper resonates with the remarkable
experimental observation of head stabilization during walking in
humans~\cite{pozzo1990head}, and ensuing locomotion models where COM and
stabilized head play complementary roles~\cite{laumond2015isrr}. Exploring this
connection is another possible direction of research.}

\begin{table}
    \centering
    \caption{Gains and weights used in the differential IK tracker \newline 
    (\textsc{n/a}: no gain for tasks regulating accelerations)}
    \label{table:ik}
    \begin{tabular}{|l|c|c|}
        \hline
        \textsc{Task description} & \textsc{Gain} [Hz] & \textsc{Weight} \\ \hline \hline
        Contacting end-effector & 1 & 100 \\ \hline
        Free end-effector & 0.03 & 1 \\ \hline
        Center of mass tracking & 1 & 5 \\ \hline
        Angular momentum variations & \textsc{n/a} & 0.2 \\ \hline
        \ins{Joint-limit gain} $K_s$ & 50 & \textsc{n/a} \\ \hline
        Velocity smoothness & \textsc{n/a} & 1 \\ \hline
        Preferred joint-angles & 0.05 & 0.1 \\ \hline
    \end{tabular}
\end{table}

\begin{table}
    \centering
    \caption{Simulation and trajectory generation parameters}
    \label{table:sims}
    \begin{tabular}{|l|c|c|}
        \hline
        \textsc{Description} & \textsc{Symbol} & \textsc{Value} \\
        \hline
        \hline
        Friction coefficient (all contacts) & $\mu$ & 0.5 \\ \hline
        Number of traj. gen. timesteps & $K$ & 100 \\ \hline
        Duration of traj. gen. timesteps & $\delta t$ & 10 ms \\ \hline
        Plane normal & $\bfn$ & $[0\ 0\ 1]$ \\ \hline
        Step duration & $T_S$ & 2.5 [s] \\ \hline
        Velocity limits & $\dot{\bfq}_{\textrm{max}}$ & 0.5 [rad/s] \\ \hline
    \end{tabular}
\end{table}

\section*{Acknowledgment}

This research was supported by 2014-2015 NEDO International R\&D and
Demonstrative Project in Environmental and Medical Fields / International R\&D
and Demonstrative Project in Robotics Fields / "Research and Development of
Robot Open Platform for Disaster Response" (PI: Yoshihiko Nakamura).

\bibliographystyle{IEEEtran}
\bibliography{refs}

\appendices

\section{Perspectives on a three-dimensional ZMP}
\label{app:better}

\ins{Considering }Proposition~\ref{prop:invar2d}\del{ implies that, while the moment
$\bftau_O$ depends on the reference point $O$, its projection $Z$ on $\Pi$ does
not depend on the planar coordinates of $O$. In this sense}, the ZMP decouples
the moment of a wrench from the position at which it is taken, yet at the
``cost'' of one dimension\del{, as it only represents two out of the three
moment coordinates}. A natural question is then: could a \del{``generalized''}
three-dimensional ZMP perform a similar decoupling for all three coordinates of
the moment? Unfortunately the answer seems to be negative, at least in the
following sense:

\begin{proposition}
    Let $\bfp_Z(O) = \bfB(\bff)\, \bftau_O$ denote any linear projection of the
    moment $\bftau_O$, where the matrix $\bfB(\bff)$ is allowed to depend
    non-linearly on $\bff$. The set of displacements $\overrightarrow{OO'}$ of
    the reference point $O$ that leave $Z$ invariant is a vector space of
    dimension at most two. 
\end{proposition}

\addtolength{\textheight}{-10cm}

\begin{IEEEproof}
    Let $O$ and $O'$ denote two points such that $\bfp_Z(O') = \bfp_Z(O)$. Then,
    $\overrightarrow{O'O} + \bfB(\bff) \overrightarrow{OO'} \times \bff
    = \bm{0}$, which rewrites to $\bfC\,\overrightarrow{OO'}
    = \overrightarrow{OO'}$ for $\bfC \defeq \bfB(\bff) [-\bff \times]$. The
    translation vector $\overrightarrow{OO'}$ thus belongs to the eigenspace
    $\calE$ of $\bfC$ associated to the eigenvalue 1. To conclude, remark that
    $\textrm{dim}(\calE) \leq~ \textrm{rank}(\bfC) \leq~\textrm{rank}([\bff
    \times]) \leq 2$.
\end{IEEEproof}

In other words, at least one coordinate of the ZMP depends on the reference
point $O$. Let us then consider the remaining moment coordinate\ins{ $(\bfn
\cdot \bftau_O)$}\del{ which is not represented by the conventional ZMP}. \del{Our
analysis following Equation~\eqref{eq:zmp} can be applied to this coordinate:}
\ins{Applying the same calculations as those following Equation~\eqref{eq:zmp},
one can write:}
\begin{equation*}
    \frac{\bfn \cdot \bftau_O}{\bfn \cdot \bff}
    \ = \ \frac{\sum_i \lambda_i \bfn \cdot \bfp_{C_i} \times
    \bff_i}{\sum_i \lambda_i (\bfn \cdot \bff_i)} \\
    \ = \ \frac{\sum_i \lambda_i (\bfn \cdot \bff_i) {\frac{\bfn \cdot
    \bfp_{C_i} \times \bff_i}{\bfn \cdot \bff_i}}}{\sum_i \lambda_i
    (\bfn \cdot \bff_i)}
\end{equation*}
\del{A definition of the }\ins{We then define a }spatial point including
all three coordinates, which we \del{will }call the \emph{$\bfn$-Moment Point}
($\bfn$-MP for short)\del{ is then}\ins{:}
\begin{equation}
    \label{nmp-def}
    \bfp_M \ = \ \frac{\bfn \times \bftau_O}{\bfn \cdot \bff}
    + \frac{\bfn \cdot \bftau_O}{\bfn \cdot \bff}\:\bfn.
\end{equation}
The vertices of its \emph{support volume} $\calV$ can be computed in the same fashion
as in Section~\ref{sec:contact} by
\begin{equation}
    \bfp_{M_i} \ = \ \frac{\bfn \times (\bfp_{C_i} \times \bff_i)}{\bfn \cdot
    \bff_i} + \frac{\bfn \cdot (\bfp_{C_i} \times \bff_i)}{\bfn \cdot \bff_i}\:\bfn.
\end{equation}
The geometric construction of support areas can also be applied
\emph{mutatis mutandis} to $\calV$:
\begin{itemize}
    \item when all virtual pressures $\bfn \cdot
\bff_i$ have the same sign, $\calV$ is the convex hull of the above vertices,
\item otherwise, it is the union of two polyhedral convex cones built on the
Minkowski difference of positive- and negative-pressure polyhedra.
\end{itemize}
A complete
implementation of this construction can be found in the
\textit{n\_moment\_point} folder of the accompanying source code~\cite{code}.


The $\bfn$-MP is a three dimensional spatial point equivalent to the moment
$\bftau_O$, in the sense that one can be computed from the other by
\begin{equation}
    \bftau_O 
    \ = \ \overrightarrow{OM} \times (\bfn \cdot \bff) \bfn + (\bfn \cdot
    \overrightarrow{OM}) (\bfn \cdot \bff) \bfn.
\end{equation}
In other words, $M$ represents the \emph{screw coordinates} of the contact
wrench along its non-central axis directed by $\bfn$, with magnitude $\bfn
\cdot \bff$ and pitch $\bfn \cdot \overrightarrow{OM}$.

However, adding the third moment coordinates makes the shape of the support
volume $\calV$ depend on the choice of the reference point $O$. Formally:

\begin{proposition}
    There is no non-empty subspace of displacements $\overrightarrow{OO'}$ of
    the reference point $O$, independent from the resultant $\bff$, that
    leaves the $\bfn$-MP invariant.
\end{proposition}

\begin{IEEEproof}
    Consider a displacement $\overrightarrow{OO'}$ of $O$ in the plane. From
    Equation~\eqref{nmp-def}, it results in a variation $\overrightarrow{OO'}
    \cdot \frac{\bfn \times \bff}{\bfn \cdot \bff}$ of the $\bfn$-MP coordinate
    along $\bfn$. This term needs to be zero for any displacement leaving the
    $\bfn$-MP invariant, thus $\overrightarrow{OO'}$ is parallel to either
    $\bfn$ or $\bff$. The former  would yield a variation of the plane
    coordinates of the $\bfn$-MP (that is to say, of the ZMP). The latter is
    excluded as we are looking for an invariance that is independent from the
    resultant force.
\end{IEEEproof}

\begin{IEEEbiography}[{\includegraphics[width=1in,height=1.25in,clip,keepaspectratio]{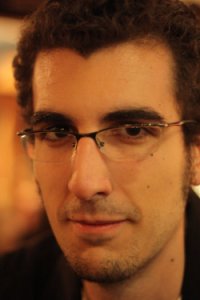}}]{St\'ephane Caron}
    is a post-doctoral researcher at the Laboratoire d’Informatique, de
    Robotique et de Micro\'{e}lectronique de Montpellier (LIRMM). Born in
    Toulouse, France, in 1988, he received the B.S and M.S. degrees from the
    \'{E}cole Normale Sup\'{e}rieure (45 rue d'Ulm), Paris, before moving to
    Japan for his doctoral studies. He received the Ph.D. in 2016 from the
    University of Tokyo, Japan. His dissertation was on whole-body motion
    planning for humanoid robots, with a focus on multi-contact stability.
\end{IEEEbiography}

\begin{IEEEbiography}[{\includegraphics[width=1in,height=1.25in,clip,keepaspectratio]{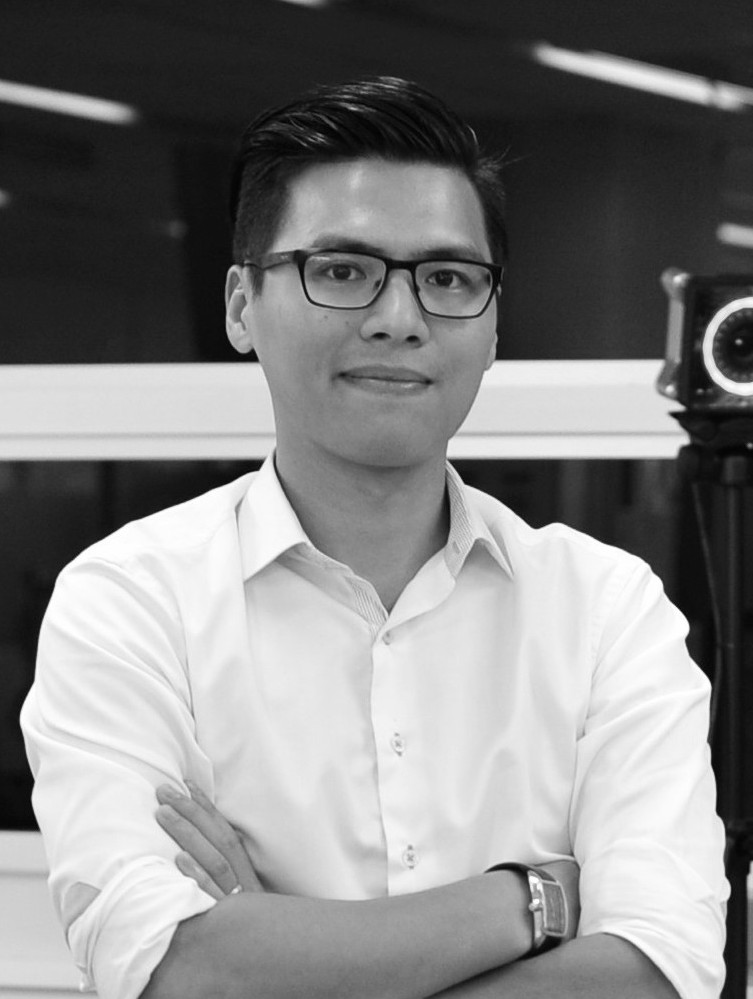}}]{Quang-Cuong Pham}
    was born in Hanoi, Vietnam. He graduated from the Departments of Computer
    Science and Cognitive Sciences of \'{E}cole Normale Sup\'{e}rieure rue
    d'Ulm, Paris, France, in 2007. He obtained a PhD in Neuroscience from
    Université Paris VI and Coll\`{e}ge de France in 2009. In 2010, he was a
    visiting researcher at the University of São Paulo, Brazil. From 2011 to
    2013, he was a researcher at the University of Tokyo, supported by a
    fellowship from the Japan Society for the Promotion of Science (JSPS). He
    joined the School of Mechanical and Aerospace Engineering, NTU, Singapore,
    as an Assistant Professor in 2013.
\end{IEEEbiography}

\begin{IEEEbiography}[{\includegraphics[width=1in,height=1.25in,clip,keepaspectratio]{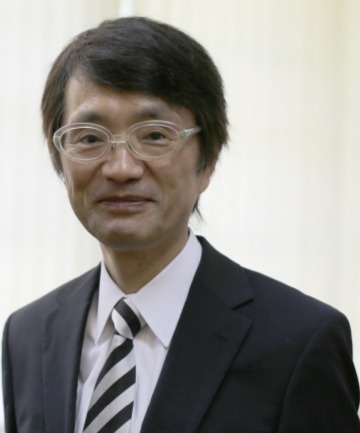}}]{Yoshihiko Nakamura}
    is Professor at the Department of Mechano-Informatics, the University of
    Tokyo, Japan. He received the Ph.D Degree from Kyoto University in 1985.
    From 1987 to 1991, he worked as Assistant and Associate Professor at the
    University of California, Santa Barbara. Humanoid robotics, cognitive
    robotics, neuro-musculoskeletal human modeling, biomedical systems, and
    their computational algorithms are his current fields of research. He is
    Fellow of JSME, Fellow of RSJ, Fellow of IEEE, and Fellow of WAAS. Dr.
    Nakamura served as President of IFToMM (2012-2015). He is a Foreign Member
    of the Academy of Engineering Science of Serbia, and TUM Distinguished
    Affiliated Professor of the Technische Universitat Munchen.
\end{IEEEbiography}

\end{document}